\documentclass[letterpaper]{article}
\let\AAAIpdfinfo\pdfinfo
\usepackage[preprint]{aaai2027}
\usepackage[hyphens]{url}
\usepackage{graphicx}
\usepackage{natbib}
\usepackage{caption}
\usepackage{booktabs}
\usepackage{amsmath,amssymb}
\usepackage{multirow}
\usepackage{array}
\usepackage{tabularx}
\graphicspath{{./}}

\newcommand{\tvadj}{\mathrm{TV}_{\mathrm{adj}}}

\providecommand{\mathbbm}[1]{\mathbb{#1}}
\newcommand{\arxivextendedconclusion}{By making attribute movement explicit,
the audit separates deployment compatibility from image-level plausibility and
localizes categories requiring correction. Its staged design connects low-cost
screening, confirmatory inference, and a bounded offline response, supporting
joint decisions about speed, quality, and semantic preservation under explicit
deployment constraints and audit budgets.}
\newcolumntype{Y}{>{\raggedright\arraybackslash}X}

\let\pdfinfo\AAAIpdfinfo
\title{DefaultShift: Auditing Semantic Default Shift in Accelerated Text-to-Image Models}
\author{Xuanhua Yin, Chuanzhi Xu, Shunqi Mao, Wei Guo, Weidong Cai\corresponding}
\affiliations{
School of Computer Science, The University of Sydney\\
\{xuanhua.yin, chuanzhi.xu, shunqi.mao, wei.guo, tom.cai\}@sydney.edu.au
}

\begin{document}
\maketitle

\begin{abstract}
Few-step text-to-image models increasingly replace slower generators, yet
acceleration can silently change distributions over unspecified attributes even
when individual outputs remain plausible and aligned. We call these
distributions \emph{semantic defaults} and their change under replacement
\emph{semantic default shift}. Existing quality, preference, and diversity
evaluations do not test whether a replacement preserves its reference model's
semantic defaults. We introduce DefaultShift, a paired audit that labels
repeated samples with closed semantic vocabularies, measures probability-mass
movement, and separates interpretable ranking from confirmatory cross-fit
inference. Across 14 reference and replacement pairs, adjusted color
discrepancies range from $0.054$ to $0.303$ with recipe-specific directions. A
1,000-image human audit reproduces the ordering. We further
introduce DefaultShift-Select, an offline calibration method that reduces
human-measured shift by $10.3\%$--$35.1\%$ across Turbo, DMD2, and FLUX
without material quality loss. Under balanced evaluation, selected data recover
$4.3$ accuracy points and $7.5$ worst-group points over uncalibrated
replacement data. DefaultShift makes semantic preservation under acceleration
measurable and actionable.
\end{abstract}

\section{Introduction}

\begin{figure}[!t]
    \centering
    \includegraphics[width=0.95\columnwidth]{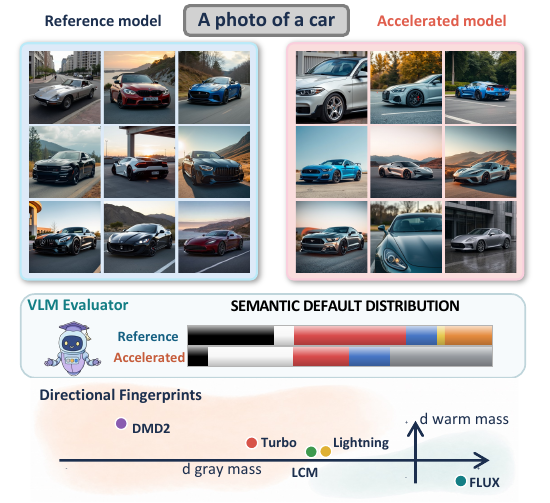}
    \caption{The same prompt can induce different semantic defaults.
    A reference model and its accelerated replacement can produce individually
    plausible outputs while following different semantic distributions.
    DefaultShift measures both the magnitude and direction of this change.}
    \label{fig:teaser}
\end{figure}

Few-step text-to-image models are increasingly deployed as replacements for
slower generators~\cite{luo2023lcm,sauer2023turbo,lin2024lightning,yin2024dmd2}.
Their intended benefit is computational: they should reduce inference cost
without changing how the system responds to the same prompt. However, such a
change can occur even when individual outputs remain realistic and
prompt-aligned. As shown in Fig.~\ref{fig:teaser}, a reference model and its
accelerated replacement can both generate plausible cars while favoring
different colors left unspecified by the prompt. No individual image is
incorrect; the difference emerges in the distribution across repeated samples.
We call these prompt-conditioned distributions over unspecified attributes a
model's \emph{semantic defaults}, and their change under replacement a
\emph{semantic default shift}.
This matters whenever accelerated outputs replace reference data in downstream
training or audited deployments~\cite{wang2023diffusiondb,luccioni2023stablebias,
struppek2023biasedartist,bianchi2023stereotypes}.

The audited attributes may include color, background, lighting, and viewpoint.
The reference defines the behavior being compared; it is not treated as a
ground-truth or ideal distribution.

Current evaluation protocols do not directly test whether these distributions
are preserved under model replacement. Quality, alignment, and preference
measures assess individual outputs~\cite{hessel2021clipscore,kirstain2023pickscore,
xu2023imagereward,hu2023tifa,ghosh2023geneval,lin2024vqascore}, while diversity
and coverage measures summarize variation across an output
set~\cite{sajjadi2018precision,kynkaanniemi2019precision,zhang2018lpips,
friedman2023vendi}. Attribute-level studies such as GRADE and DIMCIM
characterize default modes within individual models~\cite{rassin2024grade,
teotia2025dimcim}, but they are
not designed to compare a declared reference--replacement pair or track how
probability mass moves between semantic categories. Model replacement is
therefore a pairwise preservation problem and requires a reference-relative
audit.

We introduce \emph{DefaultShift} to provide this audit. For each prompt, it
draws repeated samples from the reference and replacement models, assigns each
image a label from a closed semantic vocabulary, and compares the resulting
attribute distributions. This comparison captures both the magnitude and
direction of probability-mass movement. DefaultShift also accounts for
finite-sample effects, which can make two empirical distributions appear
different even when their underlying distributions are similar. A bias-reduced
score supports interpretable ranking, while signed cross-fit estimation
supports confirmatory inference.

Across 14 reference and replacement pairs, adjusted color discrepancies range
from $0.054$ to $0.303$. The shifts depend on the acceleration recipe rather
than following a uniform pattern. DMD2 and Turbo reduce gray mass and increase
warm-color mass, whereas FLUX moves in the opposite direction. Changes in
quality, preference, and global diversity do not reliably characterize these
differences. The main ordering remains stable across evaluator swaps, prompt
and configuration stress tests, multi-attribute controls, and human
annotations. To test whether the audit can guide a practical response, we further introduce
\emph{DefaultShift-Select}, a reference-guided offline calibration method. It
selects a quality-filtered subset from a larger replacement-model candidate
pool so that the selected semantic distribution better matches the reference.
Across Turbo, DMD2, and FLUX, it reduces human-measured shift by
$10.3\%$--$35.1\%$ without material quality loss. In a balanced synthetic-data
classification study, the selected data recover $4.3$ accuracy points and
$7.5$ worst-group points over uncalibrated replacement data. Because the method
requires additional candidate generation, it is an offline calibration rather
than a cost-free online correction.

Our contributions are summarized as follows:

\begin{itemize}
    \item We introduce DefaultShift, a paired and reference-relative audit of
    semantic default shift with finite-sample-aware inference.

    \item Across 14 replacement pairs, we reveal recipe-specific and sometimes
    opposing shifts that standard quality and diversity measures do not
    reliably describe.

    \item We propose DefaultShift-Select, which reduces human-measured shift
    without material quality loss and improves downstream performance under
    balanced evaluation.
\end{itemize}

\section{Related Work}

\subsection{Efficient Text-to-Image Generation}
Fast generation combines numerical solvers~\cite{karras2022edm,lu2022dpmsolver,lu2023dpmsolverpp,zhang2023deis}, consistency or progressive distillation~\cite{salimans2022progressive,song2023consistency,meng2023guided,luo2023lcm}, adversarial and distribution matching~\cite{sauer2023turbo,lin2024lightning,yin2024dmd,yin2024dmd2}, and flow or trajectory compression~\cite{liu2023rectifiedflow,liu2024instaflow,luo2024sim,salimans2024moment,nguyen2024swiftbrush,luo2025tdm,starodubcev2026swd,ge2026senseflow}. These recipes optimize fidelity and latency, but need not preserve the same conditional behavior. Their lineage also defines the valid comparison. NitroSD-Realism, for example, is distilled from DMD2 rather than directly from SDXL~\cite{chen2025nitrofusion}. DefaultShift audits each declared deployment replacement instead of treating all fast models as equivalent.

\subsection{Generative Model Evaluation}
Aggregate fidelity and coverage metrics~\cite{heusel2017fid,sajjadi2018precision,kynkaanniemi2019precision,zhang2018lpips,oquab2024dinov2,friedman2023vendi,jayasumana2024rethinking} complement conditional alignment and preference evaluators~\cite{radford2021clip,hessel2021clipscore,kirstain2023pickscore,xu2023imagereward,wu2023hps,hu2023tifa,huang2023compbench,ghosh2023geneval,cho2024dsg,lin2024vqascore}. GRADE and DIMCIM most closely expose semantic defaults by measuring attribute diversity within one model~\cite{rassin2024grade,teotia2025dimcim}. DefaultShift instead tests preservation between a declared reference and replacement, retains category identity and direction, and quantifies finite-sample uncertainty. This paired view complements known diversity loss in few-step generators~\cite{gandikota2025distilling,xie2024em,salimans2024moment} and demographic audits~\cite{luccioni2023stablebias,struppek2023biasedartist,bianchi2023stereotypes}.

\subsection{Reference-Based Selection}
Reward learning and preference optimization improve individual outputs by updating model parameters against scalar objectives~\cite{fan2023dpok,black2024ddpo,wallace2024diffusiondpo}. Preference evaluators can also rank candidates one image at a time~\cite{kirstain2023pickscore,xu2023imagereward,wu2023hps}, but independent ranking does not constrain the joint semantic composition of the released set. DefaultShift-Select instead formulates selection as set-level distribution matching. It estimates a target histogram from reference samples, applies a frozen quality floor to a shared replacement pool, and selects a subset that minimizes distributional discrepancy against held-out reference labels. It therefore changes neither the generator nor its scalar reward. The required reference and additional candidates make it a bounded offline calibration rather than a teacher-free online method.

\section{Methodology}
\subsection{Task Formulation}

\begin{figure*}[t]
    \centering
    \includegraphics[width=\textwidth]{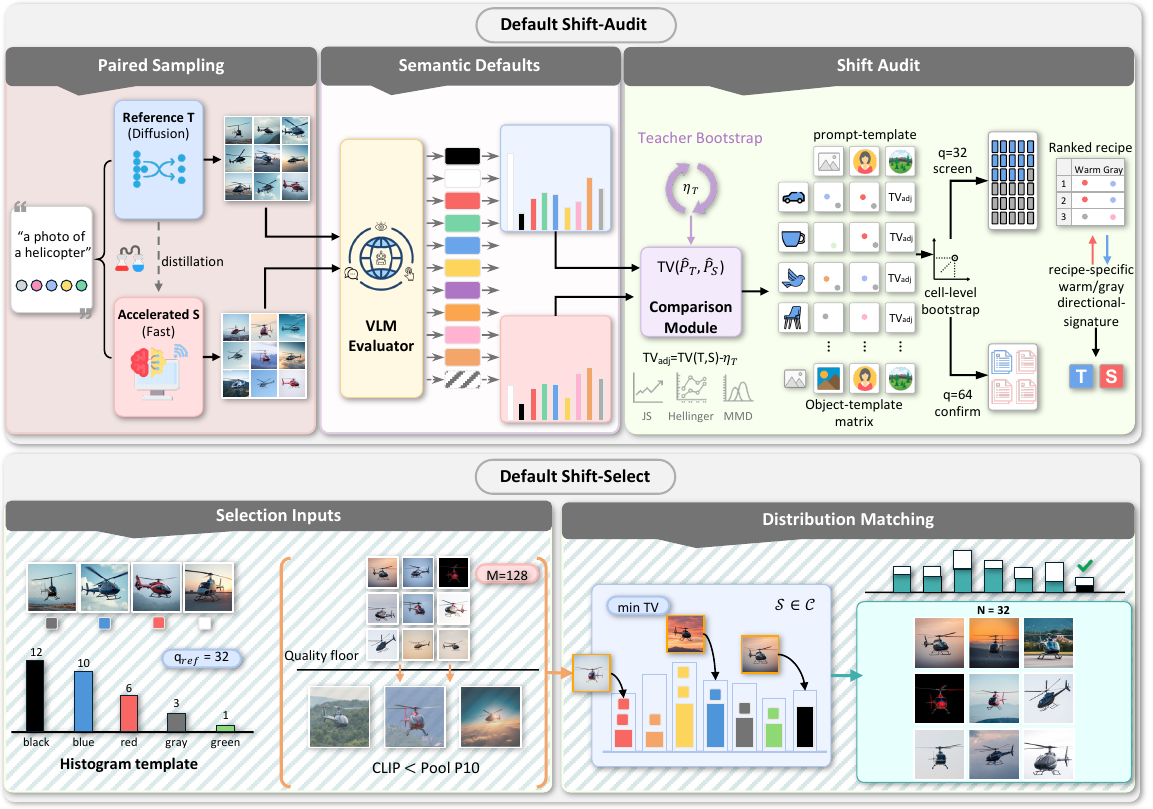}
    \caption{Overview of DefaultShift. The audit generates matched
    samples from a declared reference and accelerated replacement, maps each
    image to a closed semantic vocabulary, and compares their conditional
    distributions. Cell-level discrepancies are aggregated for recipe ranking
    and directional analysis. DefaultShift-Select uses reference statistics
    and a quality-filtered replacement pool to construct a distribution-matched
    offline subset.}
    \label{fig:defaultshift_method}
\end{figure*}

Let $T$ be a slow reference generator, $S$ an accelerated replacement, $p$ an underspecified prompt, and $a$ a semantic attribute with vocabulary $\mathcal{K}_a$. Repeated generation and attribute labeling produce empirical conditional distributions $\widehat P_T(a\mid p)$ and $\widehat P_S(a\mid p)$, which define the semantic defaults for the declared deployment pair. We audit whether replacement preserves this conditional behavior. The reference is a behavioral baseline, not an assumption of correctness, fairness, or real-world representativeness. Figure~\ref{fig:defaultshift_method} summarizes the audit and its reference-guided offline selection extension.

\subsection{Attribute Measurement}
We generate matched images whenever supported and label them with
Qwen2.5-VL-3B-Instruct~\cite{bai2025qwen25vl}; BLIP-2, LLaVA, and Qwen-VL
provide evaluator-swap checks~\cite{li2023blip2,liu2023llava,bai2023qwenvl}.
For color, the evaluator must return one of 11 labels or \emph{unclear}.
The vocabulary is black, white, red, green, blue, yellow, purple, orange,
pink, brown, and gray. We use a frozen sequential protocol. Unclear,
unparsable, and safety-blocked images receive an unknown label: they remain in
coverage statistics but not histogram normalization.

For model $m\in\{T,S\}$, let $y^i_{m,a}$ denote the attribute label assigned to sample $i$. We normalize category counts over valid labels:
\begin{equation}
\widehat P_m(a=k\mid p)=
\frac{\sum_{i=1}^{N}\mathbbm{1}[y^i_{m,a}=k]}
{\sum_{i=1}^{N}\mathbbm{1}[y^i_{m,a}\neq\bot]},
\quad k\in\mathcal K_a.
\end{equation}

\subsection{Bias-Reduced Distribution Distance}
The raw total variation for a prompt cell is: 
\begin{equation}
\begin{aligned}
\mathrm{TV}(T,S,p,a)&=\frac{1}{2}\sum_{k\in\mathcal K_a}
\bigl|\widehat P_T(a=k\mid p)\\
&\hspace{6.0em}-\widehat P_S(a=k\mid p)\bigr|.
\end{aligned}
\end{equation}
Finite samples produce positive empirical TV even under equal distributions. We estimate a matched reference noise floor by repeatedly drawing two size-$N$ bootstrap multisets:
\begin{equation}
\eta_T(p,a)=\mathbb E_b\left[
\mathrm{TV}\left(\widehat P_{T,1}^{(b)},\widehat P_{T,2}^{(b)}\right)
\right].
\end{equation}
The primary audit score is: 
\begin{equation}
\tvadj(T,S,p,a)=\mathrm{TV}(T,S,p,a)-\eta_T(p,a).
\label{eq:tvadj}
\end{equation}
We retain negative values rather than clipping them. The same matched correction is applied to Jensen-Shannon divergence, Hellinger distance, and categorical MMD. To represent direction, we retain the signed category-wise movement $\Delta_{T\rightarrow S}(k\mid p,a)=\widehat P_S(a=k\mid p)-\widehat P_T(a=k\mid p)$, where positive and negative values denote probability-mass gain and loss under replacement.

Equation~\ref{eq:tvadj} is a bias-reduced audit score, not an unbiased estimator of population TV. Reference-to-reference sampling bias need not equal reference-to-replacement sampling bias. We therefore use $q=32$ for uniform-cost screening and require $q=64$ plus a two-fold signed cross-fit sensitivity for confirmatory effect-size claims. Cross-fit learns the sign of each category contrast on one sample half, evaluates the signed contrast on the other half, swaps the halves, and averages. Frozen multinomial simulations quantify bias, RMSE, interval coverage, and ranking recovery across $q\in\{8,16,32,64,96\}$.

\subsection{Aggregation and Statistical Unit}
An object--template pair is one audit cell, and recipe means weight 192 cells
equally. The $q=64$ endpoint contains 12,288 labels per model-side bank.
Intervals use 2,000 object-cluster bootstraps over the 48 objects; held-out
selection comparisons use 5,000 repetitions. Unknown-label sensitivities either
retain unknown as a category or adjust by valid-label coverage. We froze seeds,
VLM questions, configurations, endpoints, and failure rules before expansion.

We use $\tvadj$ to rank recipes and localize category movement. Confirmatory
claims require $q=64$, signed cross-fit, and consistent ordering across
Jensen--Shannon, Hellinger, and categorical MMD. Mechanism probes remain
associative unless an intervention changes the proposed factor.

\subsection{DefaultShift-Select}
DefaultShift-Select is a reference-guided offline calibration. For each held-out
object, it estimates a target histogram from 32 reference labels, draws
$M\in\{32,64,128\}$ replacement candidates, and selects $N=32$ from the shared
VLM-valid pool. Let $\mathcal C_\tau$ be the candidates that pass the frozen
CLIP quality floor and let $h(\mathcal A)$ denote the normalized attribute
histogram of subset $\mathcal A$. The selection target is
\begin{equation}
\mathcal A^\star=\underset{\substack{\mathcal A\subseteq\mathcal C_\tau\\
|\mathcal A|=N}}{\arg\min}\;
\mathrm{TV}\bigl(h(\mathcal A),\widehat P_T\bigr).
\end{equation}
Greedy updates approximate this discrete objective. We compare random,
quality-only, feature-diversity, semantic-entropy, Select, and a same-pool
minimum-TV oracle. All baselines share candidate pools, quality floors, budgets,
and held-out objects. A two-fold 16/16 reference split separates target
estimation from evaluation, so reported gains use reference labels unavailable
to selection. The operation changes which replacement outputs are released
rather than generator weights or sampling dynamics.

\section{Experiments and Results}
\label{sec:experiments}

\subsection{Experimental Setup}
\label{sec:exp_setup}

\paragraph{Benchmark.}
We audit 14 reference and replacement pairs spanning consistency, progressive and
adversarial distillation, distribution matching, trajectory matching, and
flow distillation. The panel includes Hyper-SD15~\cite{ren2024hypersd},
Flash-SDXL~\cite{chadebec2025flash}, PixArt-LCM~\cite{chen2024pixartdelta},
and the FLUX.1 [schnell] checkpoint~\cite{blackforestlabs2024flux}. The unified analysis contains 48 objects and four prompt
templates, yielding 192 cells for every replacement pair. Each checkpoint is
paired with its intended deployment reference under the official scheduler,
guidance, resolution, and sampling steps. We use paired seeds when supported.
The uniform-cost screen uses $q=32$ samples per cell. The confirmatory endpoint
uses $q=64$ for all 14 replacement pairs. Historical 100-object results are
retained as replication evidence.

\paragraph{Measurement.}
Color is primary. We also measure background, lighting, and viewpoint.
Qwen2.5-VL-3B-Instruct is the primary VLM attribute evaluator, with BLIP and LLaVA swaps.
We report $\tvadj$ with adjusted Jensen--Shannon divergence, Hellinger
distance, categorical MMD, raw TV, and signed cross-fit TV as sensitivities.
Intervals use 2,000 object-cluster bootstrap repetitions unless stated
otherwise. Unknown labels remain in coverage analyses. We also compute
GRADE-style normalized attribute entropy on the identical label banks.

\paragraph{Human Validation.}
Our first study contains 300 stratified images from the reference, Turbo, and
LCM across 25 objects and four templates. Fifteen annotators provide three
independent judgments per image. Majority vote defines the human label. A
second preregistered study contains 1,000 images balanced across six reference
and replacement pairs. It tests replacement-level distribution distances,
source-stratified evaluator accuracy, and independent rank reproduction.

\subsection{Cross-Recipe Default Shift}
\label{sec:main_results}

\begin{figure*}[t]
    \centering
    \includegraphics[width=0.98\textwidth]{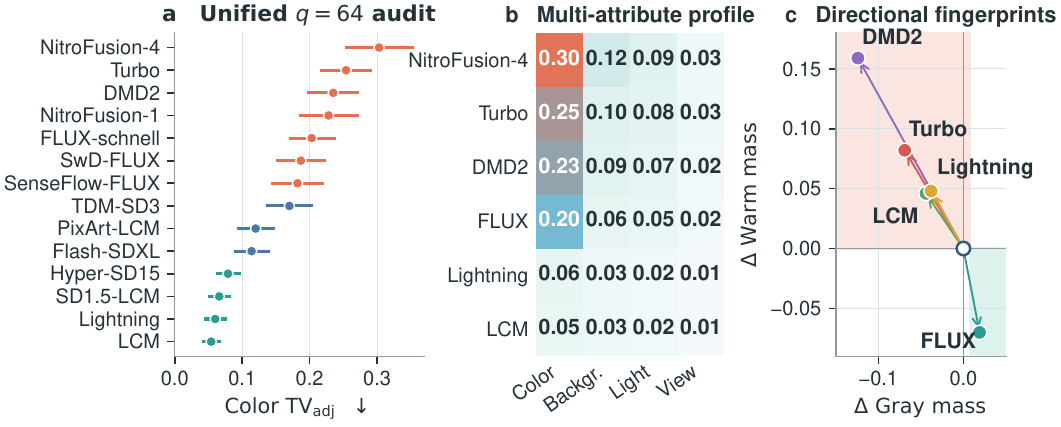}
    \caption{Acceleration recipes differ in magnitude, attribute, and
    direction. a The unified $q=64$ audit covers all 14 replacement
    pairs. Lines are 95\% object-cluster intervals. b Color dominates
    the four-attribute profile, while viewpoint shifts are small.
    c Replacement-minus-reference changes in gray and warm mass reveal
    opposing recipe fingerprints.}
    \label{fig:results_overview}
\end{figure*}

Figure~\ref{fig:results_overview}a shows a continuous recipe-dependent
spectrum rather than a binary fast versus slow effect. The unified $q=64$
endpoint ranges from $0.303$ for NitroFusion-4 to $0.054$ for LCM. Turbo and
DMD2 remain large at $0.254$ and $0.235$. The corresponding $q=32$ and $q=64$
recipe rankings have Spearman correlation $0.996$. The
correlation between $\tvadj$ and signed cross-fit TV is $0.994$. The pooled
SDXL aggressive-minus-mild contrast is $0.207$ with interval [0.171, 0.243].
Table~\ref{tab:q64confirm} reports every confirmatory estimate.

\begin{table*}[t]
    \centering
    \small
    \setlength{\tabcolsep}{1.4pt}
    \renewcommand{\arraystretch}{1.06}
    \begin{tabular}{lrrl@{\hspace{8pt}}lrrl}
        \toprule
        Recipe & $q=32$ & $q=64$ & 95\% CI & Recipe & $q=32$ & $q=64$ & 95\% CI \\
        \midrule
        NitroFusion-4 & 0.269 & \bfseries 0.303 & [0.252, 0.354] & TDM-SD3 & 0.155 & 0.170 & [0.136, 0.205] \\
        Turbo & 0.226 & \bfseries 0.254 & [0.215, 0.293] & PixArt-LCM & 0.107 & 0.120 & [0.092, 0.148] \\
        DMD2 & 0.210 & \bfseries 0.235 & [0.196, 0.273] & Flash-SDXL & 0.101 & 0.114 & [0.088, 0.141] \\
        NitroFusion-1 & 0.205 & 0.228 & [0.184, 0.273] & Hyper-SD15 & 0.073 & 0.079 & [0.061, 0.098] \\
        FLUX-schnell & 0.192 & 0.203 & [0.169, 0.239] & SD1.5-LCM & 0.060 & 0.066 & [0.050, 0.083] \\
        SwD-FLUX & 0.171 & 0.187 & [0.150, 0.225] & Lightning & 0.040 & \bfseries 0.060 & [0.044, 0.077] \\
        SenseFlow-FLUX & 0.167 & 0.182 & [0.143, 0.222] & LCM & 0.045 & \bfseries 0.054 & [0.040, 0.069] \\
        \bottomrule
    \end{tabular}
    \caption{Unified color audit. The $q=32$ screen and $q=64$
    endpoint use identical cells. Brackets give 95\% object-cluster intervals
    for the $q=64$ endpoint.}
    \label{tab:q64confirm}
\end{table*}

The discrepancy is color-led rather than uniform. At $q=64$, background
$\tvadj$ is $0.115$ [0.087, 0.145] for NitroFusion-4, $0.100$ [0.078, 0.123] for
Turbo, $0.086$ [0.066, 0.108] for DMD2, $0.060$ [0.038, 0.083] for FLUX,
$0.030$ [0.018, 0.043] for Lightning, and $0.028$ [0.015, 0.041] for LCM.
Lighting ranges from $0.022$ to $0.090$, while viewpoint ranges from $0.008$
to $0.030$. The aggressive-minus-mild contrast is positive for lighting at
$0.051$ with interval [0.030, 0.073]. The viewpoint contrast of $0.016$ has
interval [-0.001, 0.033], so we make no general viewpoint claim. Background
ranking is sensitive to abstentions at $q=32$, but stabilizes at $q=64$. Its
primary-versus-unknown Spearman correlation rises from $0.60$ to $0.83$, while
a coverage-adjusted sensitivity reaches $0.97$.

Figure~\ref{fig:results_overview}c shows that direction also matters. DMD2 and
Turbo lose gray and gain warm mass, while FLUX moves oppositely. Turbo, DMD2,
and FLUX all lie in the aggressive group, yet directional information changes
the diagnosis even when aggregate magnitudes are close. A scalar distance
alone cannot distinguish their opposing category movements. Retaining the
category-wise vector therefore identifies which modes a reference-matching
intervention should replenish or suppress. Lineage changes the interpretation
of NitroFusion-Realism. Its checkpoint is distilled from DMD2 rather than
directly from SDXL~\cite{chen2025nitrofusion}. We therefore describe SDXL to
NitroFusion-Realism as end-to-end deployment-chain drift and do not attribute the full
value to one training stage.

\subsection{Reliability and Scope}
\label{sec:validity}

\begin{figure*}[t]
    \centering
    \includegraphics[width=0.90\textwidth]{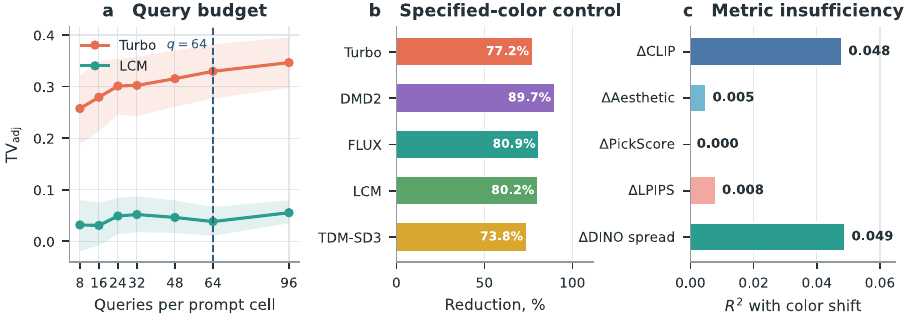}
    \caption{Reliability and specificity. a Query-budget
    convergence on a 24-object plain-prompt diagnostic, not the main endpoint.
    b Explicitly
    specifying color suppresses shift across five recipes. c
    Conventional metrics explain little object-level color shift. Bands denote
    95\% cell-bootstrap intervals.}
    \label{fig:validation_overview}
\end{figure*}

\begin{figure*}[t]
    \centering
    \includegraphics[width=0.97\textwidth]{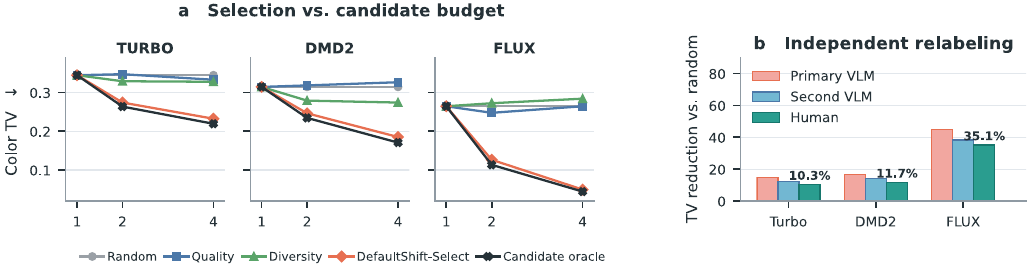}
    \caption{Reference-guided offline calibration.
    a DefaultShift-Select approaches a candidate-pool oracle as the
    budget grows. b Improvements remain positive under an independent
    evaluator and human labels.}
    \label{fig:select_overview}
\end{figure*}

The main ordering is stable across adjusted Jensen--Shannon, Hellinger, and
categorical MMD. Their rank correlations with $\tvadj$ are $0.987$, $0.974$,
and $0.934$. In contrast, changes in CLIP, aesthetic score, PickScore, LPIPS,
and DINO spread explain $0.048$, $0.005$, $0.000$, $0.008$, and $0.049$ of
object-level color-shift variance. These metrics remain useful for quality or
global diversity, but do not identify which named semantic modes moved.

A direct GRADE-style comparison is more informative. Absolute normalized
entropy change has rank correlation $0.829$ with $\tvadj$ across the six-pair
panel. This confirms substantial overlap with diversity contraction. Entropy
discards category identity, however. Among 1,152 cells, 67 have entropy change
no greater than $0.05$ but raw TV of at least $0.30$. DefaultShift therefore
adds a declared deployment reference and the direction of moved semantic mass
rather than replacing diversity measurement.

Matched-resolution controls retain $\tvadj$ values of $0.270$ for Turbo at
512 pixels and $0.250$ for DMD2 at 1024 pixels. Specifying color reduces the
discrepancy by $73.8\%$ to $89.7\%$ across five recipes. This supports the
interpretation that the endpoint responds to semantic freedom left by the
prompt. Repeated configurations give means of $0.251$, $0.229$, $0.201$,
$0.062$, and $0.056$ for Turbo, DMD2, FLUX, Lightning, and LCM. A mixed model
retains an aggressive-recipe coefficient of $0.118$ with interval [0.079,
0.157] after accounting for teacher entropy, CFG, resolution, and steps.
Resolution itself is inconclusive at $0.008$ with interval [-0.009, 0.024].
Object and template intraclass correlations are $0.184$ and $0.029$, showing
that most clustered heterogeneity is object-specific.

Prompt-suite stress tests preserve the recipe ordering. Aggressive-minus-mild
contrasts are $0.173$ for short prompts, $0.103$ for long prompts, $0.136$ for
style prompts, $0.143$ for multilingual prompts, and $0.125$ for real-user
prompts. The minimum cross-suite rank correlation is $0.87$ with interval
[0.71, 0.96]. Multilingual coverage falls to $0.79$, which limits claims for
that suite.

\begin{table}[t]
    \centering
    \small
    \setlength{\tabcolsep}{4.0pt}
    \renewcommand{\arraystretch}{1.05}
    \begin{tabular}{lccc}
        \toprule
        Estimator & Null bias & Type I error & Rank $\rho$ \\
        \midrule
        Raw TV & 0.197 & 0.980 & 0.991 \\
        $\tvadj$ & 0.008 & 0.560 & \bfseries 0.996 \\
        Signed cross-fit & \bfseries 0.001 & \bfseries 0.052 & 0.981 \\
        Dirichlet posterior & 0.034 & 0.071 & 0.989 \\
        \bottomrule
    \end{tabular}
    \caption{Estimator simulation at $q=64$. Null calibration and
    alternative ranking are computed from frozen multinomial families.}
    \label{tab:estimator_simulation}
\end{table}

Table~\ref{tab:estimator_simulation} clarifies the statistical role of
$\tvadj$. It removes most raw-TV inflation and preserves ranking, but its
bootstrap intervals are not calibrated for population TV. Signed cross-fit
attains $0.947$ null coverage and Type I error $0.052$ at $q=64$, while its
power rises from $0.746$ at $q=32$ to $0.921$ at $q=64$. We therefore use
$\tvadj$ for ranking and cross-fit inference for confirmatory claims.

\begin{table}[t]
    \centering
    \small
    \setlength{\tabcolsep}{3.7pt}
    \renewcommand{\arraystretch}{1.06}
    \begin{tabular}{lccccc}
        \toprule
        Attribute & Clear $n$ & Cov. & Acc. & Macro-F1 & Fleiss $\kappa$ \\
        \midrule
        Color & 261 & 0.87 & 0.91 & 0.91 & 0.71 \\
        Background & 246 & 0.82 & 0.87 & 0.87 & 0.64 \\
        \bottomrule
    \end{tabular}
    \caption{Independent human validation. Each attribute is judged
    on 300 images. Clear $n$ excludes abstentions, while coverage retains them
    in the denominator. Accuracy and macro-F1 compare Qwen with majority vote.}
    \label{tab:human_validation}
\end{table}

\paragraph{Evaluator and Human Validation.}
Table~\ref{tab:human_validation} reports the completed independent study.
Conservative accuracy is $0.79$ for color and $0.71$ for background. Fleiss
$\kappa$ is $0.71$ [0.65, 0.77] and $0.64$ [0.57, 0.71], with Krippendorff
$\alpha$ of $0.70$ and $0.63$. Pairwise source comparisons show no labeling
penalty. Grouped human labels also reproduce the
distributional result. Bias-corrected color TV is $0.18$ [0.09, 0.27] for
Turbo and $0.10$ [0.02, 0.19] for LCM, with paired difference $0.08$
[0.03, 0.13]. Background yields $0.13$ [0.05, 0.21] and $0.07$ [0.00, 0.14].
Its difference is $0.06$ [-0.01, 0.13] with permutation $p=0.09$, so this
contrast is inconclusive.

\begin{figure}[t]
    \centering
    \includegraphics[width=0.92\columnwidth]{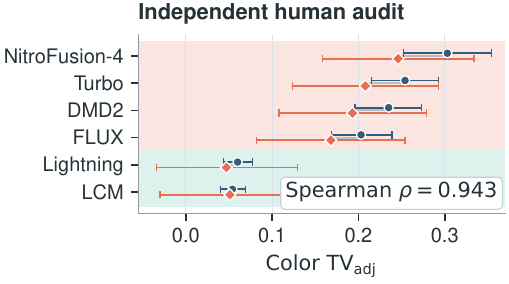}
    \caption{Independent human labels reproduce recipe separation.
    VLM circles and human diamonds show 95\% intervals. Ordering is compared
    because label granularity differs.}
    \label{fig:human_alignment}
\end{figure}

The 1,000-image expansion tests whether this evidence generalizes beyond two
replacements. Overall coverage is $0.862$, clear-case accuracy is $0.901$,
macro-F1 is $0.874$, Fleiss $\kappa$ is $0.698$, and Krippendorff $\alpha$ is
$0.68$. The aggressive-minus-reference accuracy difference is $-0.031$ with
interval [-0.079, 0.017], providing no evidence of source-dependent error.
Figure~\ref{fig:human_alignment} shows that human estimates separate all four
aggressive replacements from the two mild replacements. Human and VLM rankings
have Spearman correlation $0.943$ with interval [0.543, 1.000].

\subsection{Decomposition and Actionability}
\label{sec:calibration}

On the separate 100-object, four-template, four-attribute $q=32$ decomposition
panel, Turbo's mean $\tvadj$ decreases from $0.093$ to $0.026$
when semantic entropy is preserved and $0.013$ when both entropy and Vendi
diversity are preserved. Much of the aggregate shift therefore overlaps known coverage
degradation. The entropy-preserved subset retains an adjusted mode-flip rate
of $0.054$ with interval [0.036, 0.074]. In separate $q=32$ teacher-first
probes over 20 objects and four templates, the diversity hybrid reduces DMD2
from $0.201$ to $0.141$ and Turbo at 512 pixels from $0.297$ to
$0.160$~\cite{gandikota2025distilling}. We interpret DefaultShift as a semantic
decomposition of acceleration-related distribution change, not as a failure
mode independent of diversity degradation.

DMD2's direction has cosine similarity $0.923$ with increasing teacher CFG and
an effective CFG near $15$. Turbo has cosine similarity $0.269$ with this axis.
Its shift instead correlates with teacher color entropy at Spearman $0.596$ and
increases peak-mode probability by $0.194$. These probes distinguish recipes,
but do not establish complete causal identification.

At a fourfold candidate budget, DefaultShift-Select reduces primary-evaluator
TV by $14.8\%$, $16.8\%$, and $44.8\%$ for Turbo, DMD2, and FLUX. Independent
VLM reductions are $12.1\%$, $13.9\%$, and $38.2\%$. Human reductions are
$10.3\%$, $11.7\%$, and $35.1\%$. Their intervals are [2.9, 17.8], [3.8,
20.6], and [24.3, 46.2], respectively.
Integer quota matching reaches $16.2\%$, $18.9\%$, and $46.7\%$, defining a
strong discrete-matching baseline. Quality-only, DINO diversity, semantic
entropy, and rejection sampling are weaker on every pair.

Noninferiority tests find no material quality loss. The CLIP difference is
$0.001$ with interval [-0.003, 0.005] under margin $-0.010$. Aesthetic,
PickScore, human quality, within-color DINO diversity, and Vendi intervals also
remain above their frozen margins. Replacement inference costs $0.148$ of the
reference. Fourfold selection costs $0.652$, preserving a $1.53$-fold speedup.
The method is therefore a reference-based offline calibration, not a free
online repair.

\begin{table}[t]
    \centering
    \small
    \setlength{\tabcolsep}{2.0pt}
    \renewcommand{\arraystretch}{1.05}
    \begin{tabular}{lcc}
        \toprule
        Training source & Accuracy & Worst group \\
        \midrule
        Reference & 78.4 [76.9, 79.8] & 70.1 [67.2, 73.0] \\
        Replacement & 72.1 [70.5, 73.6] & 59.8 [56.5, 63.1] \\
        Quality-only & 72.6 & 61.2 \\
        DefaultShift-Select & \bfseries 76.4 [74.9, 77.9] & \bfseries 67.3 [64.2, 70.3] \\
        Teacher-first hybrid & 76.9 & 68.2 \\
        \bottomrule
    \end{tabular}
    \caption{Downstream classifier under balanced evaluation.
    Intervals are bootstrap 95\% intervals.}
    \label{tab:downstream}
\end{table}

Table~\ref{tab:downstream} tests why semantic preservation matters beyond the
audit score. A classifier trained on accelerated-model synthetic data loses
$6.3$ accuracy points and $10.3$ worst-group points relative to reference data.
DefaultShift-Select recovers $4.3$ accuracy points with interval [2.2, 6.4]
and $7.5$ worst-group points with interval [3.4, 11.6]. On a natural-frequency
test set, its gain is $1.1$ points. Its interval extends from $-0.8$ to $3.0$.
The benefit is
therefore concentrated where evaluation exposes the semantic modes that the
accelerated generator underrepresents.

Together, the results support a staged audit. A $q=32$ pass prioritizes pairs
under a common budget. A $q=64$ rerun with signed cross-fit inference supports
effect-size claims, while directional histograms localize the moved categories.
The protocol tests preservation rather than fairness. Evidence is strongest
for color and balanced downstream evaluation. Background effects are smaller,
viewpoint is inconclusive, multilingual coverage is lower, and Select is tested
on three models.

\section{Conclusion}
DefaultShift treats accelerated text-to-image deployment as a paired semantic
preservation problem. It compares prompt-conditioned attribute distributions
between a declared reference and replacement, reduces finite-sample inflation,
and reports shift magnitude and direction. Across 14 pairs, the confirmatory
audit reveals color-led and recipe-specific shifts that changes in quality,
preference, and global diversity do not reliably characterize. Human
validation, simulations, and stress tests support the ordering.
DefaultShift-Select reduces human-measured shift
across Turbo, DMD2, and FLUX while preserving
quality and improving balanced downstream performance. \arxivextendedconclusion\ Evidence is strongest
for color. Background shifts are smaller, viewpoint is inconclusive,
multilingual coverage is lower, and selection covers three models. DefaultShift
audits declared behavioral baselines rather than social ground truth, while its
mechanism probes remain associative rather than causal accounts.

\clearpage
\bibliography{defaultshift}

@article{luo2023lcm,
  title={Latent Consistency Models: Synthesizing High-Resolution Images with Few-Step Inference},
  author={Luo, Simian and Tan, Yiqin and Huang, Longbo and Li, Jian and Zhao, Hang},
  journal={arXiv preprint arXiv:2310.04378},
  year={2023}
}

@inproceedings{sauer2023turbo,
  title={Adversarial Diffusion Distillation},
  author={Sauer, Axel and Lorenz, Dominik and Blattmann, Andreas and Rombach, Robin},
  booktitle={European Conference on Computer Vision (ECCV)},
  year={2024}
}

@article{lin2024lightning,
  title={{SDXL-Lightning}: Progressive Adversarial Diffusion Distillation},
  author={Lin, Shanchuan and Wang, Anran and Yang, Xiao},
  journal={arXiv preprint arXiv:2402.13929},
  year={2024}
}

@inproceedings{yin2024dmd2,
  title={Improved Distribution Matching Distillation for Fast Image Synthesis},
  author={Yin, Tianwei and
          Gharbi, Micha{\"e}l and
          Park, Taesung and
          Zhang, Richard and
          Shechtman, Eli and
          Durand, Fr{\'e}do and
          Freeman, William T.},
  booktitle={Advances in Neural Information Processing Systems (NeurIPS)},
  year={2024}
}

@inproceedings{luo2025tdm,
  title={Learning Few-Step Diffusion Models by Trajectory Distribution Matching},
  author={Luo, Yihong and Hu, Tianyang and Sun, Jiacheng and Cai, Yujun and Tang, Jing},
  booktitle={IEEE/CVF International Conference on Computer Vision (ICCV)},
  pages={17719--17728},
  year={2025}
}

@inproceedings{chen2025nitrofusion,
  title={NitroFusion: High-Fidelity Single-Step Diffusion through Dynamic Adversarial Training},
  author={Chen, Dar-Yen and Bandyopadhyay, Hmrishav and Zou, Kai and Song, Yi-Zhe},
  booktitle={IEEE/CVF Conference on Computer Vision and Pattern Recognition (CVPR)},
  year={2025}
}

@inproceedings{gandikota2025distilling,
  title={Distilling Diversity and Control in Diffusion Models},
  author={Gandikota, Rohit and Bau, David},
  booktitle={IEEE/CVF Winter Conference on Applications of Computer Vision (WACV)},
  pages={1304--1313},
  year={2026}
}

@inproceedings{starodubcev2026swd,
  title={Scale-wise Distillation of Diffusion Models},
  author={Starodubcev, Nikita and Drobyshevskiy, Ilya and Kuznedelev, Denis and Babenko, Artem and Baranchuk, Dmitry},
  booktitle={International Conference on Learning Representations (ICLR)},
  year={2026}
}

@inproceedings{ge2026senseflow,
  title={SenseFlow: Scaling Distribution Matching for Flow-based Text-to-Image Distillation},
  author={Ge, Xingtong and Zhang, Xin and Xu, Tongda and Zhang, Yi and Zhang, Xinjie and Wang, Yan and Zhang, Jun},
  booktitle={International Conference on Learning Representations (ICLR)},
  year={2026}
}

@inproceedings{radford2021clip,
  title={Learning Transferable Visual Models From Natural Language Supervision},
  author={Radford, Alec and others},
  booktitle={International Conference on Machine Learning (ICML)},
  pages={8748--8763},
  year={2021}
}

@inproceedings{kirstain2023pickscore,
  title={{Pick-a-Pic}: An Open Dataset of User Preferences for Text-to-Image Generation},
  author={Kirstain, Yuval and Polyak, Adam and Singer, Uriel and Matiana, Shahbuland and Penna, Joe and Levy, Omer},
  booktitle={Advances in Neural Information Processing Systems (NeurIPS)},
  year={2023}
}

@article{friedman2023vendi,
  title={The Vendi Score: A Diversity Evaluation Metric for Machine Learning},
  author={Friedman, Dan and Dieng, Adji Bousso},
  journal={Transactions on Machine Learning Research},
  year={2023}
}

@inproceedings{liu2023rectifiedflow,
  title={Flow Straight and Fast: Learning to Generate and Transfer Data with Rectified Flow},
  author={Liu, Xingchao and Gong, Chengyue and Liu, Qiang},
  booktitle={International Conference on Learning Representations (ICLR)}, year={2023}
}

@inproceedings{salimans2022progressive,
  title={Progressive Distillation for Fast Sampling of Diffusion Models},
  author={Salimans, Tim and Ho, Jonathan},
  booktitle={International Conference on Learning Representations (ICLR)}, year={2022}
}

@inproceedings{song2023consistency,
  title={Consistency Models},
  author={Song, Yang and Dhariwal, Prafulla and Chen, Mark and Sutskever, Ilya},
  booktitle={International Conference on Machine Learning (ICML)}, year={2023}
}

@inproceedings{meng2023guided,
  title={On Distillation of Guided Diffusion Models},
  author={Meng, Chenlin and Rombach, Robin and Gao, Ruiqi and Kingma, Diederik and Ermon, Stefano and Ho, Jonathan and Salimans, Tim},
  booktitle={IEEE/CVF Conference on Computer Vision and Pattern Recognition (CVPR)}, year={2023}
}

@inproceedings{yin2024dmd,
  title={One-Step Diffusion with Distribution Matching Distillation},
  author={Yin, Tianwei and Gharbi, Micha{\"e}l and Zhang, Richard and Shechtman, Eli and Durand, Fr{\'e}do and Freeman, William T. and Park, Taesung},
  booktitle={IEEE/CVF Conference on Computer Vision and Pattern Recognition (CVPR)}, year={2024}
}

@inproceedings{nguyen2024swiftbrush,
  title={SwiftBrush: One-Step Text-to-Image Diffusion Model with Variational Score Distillation},
  author={Nguyen, Thuan Hoang and Tran, Anh},
  booktitle={IEEE/CVF Conference on Computer Vision and Pattern Recognition (CVPR)}, year={2024}
}

@inproceedings{liu2024instaflow,
  title={InstaFlow: One Step is Enough for High-Quality Diffusion-Based Text-to-Image Generation},
  author={Liu, Xingchao and Zhang, Xiwen and Ma, Jianzhu and Peng, Jian and Liu, Qiang},
  booktitle={International Conference on Learning Representations (ICLR)}, year={2024}
}

@inproceedings{xie2024em,
  title={{EM} Distillation for One-Step Diffusion Models},
  author={Xie, Sirui and Xiao, Zhisheng and Kingma, Diederik P. and Hou, Tingbo and Wu, Ying Nian and Murphy, Kevin and Salimans, Tim and Poole, Ben and Gao, Ruiqi},
  booktitle={Advances in Neural Information Processing Systems (NeurIPS)}, year={2024}
}

@inproceedings{luo2024sim,
  title={One-Step Diffusion Distillation through Score Implicit Matching},
  author={Luo, Weijian and Huang, Zemin and Geng, Zhengyang and Kolter, J. Zico and Qi, Guo-Jun},
  booktitle={Advances in Neural Information Processing Systems (NeurIPS)}, year={2024}
}

@inproceedings{salimans2024moment,
  title={Multistep Distillation of Diffusion Models via Moment Matching},
  author={Salimans, Tim and Mensink, Thomas and Heek, Jonathan and Hoogeboom, Emiel},
  booktitle={Advances in Neural Information Processing Systems (NeurIPS)}, year={2024}
}

@inproceedings{chadebec2025flash,
  title={Flash Diffusion: Accelerating Any Conditional Diffusion Model for Few Steps Image Generation},
  author={Chadebec, Cl{\'e}ment and Tasar, Onur and Benaroche, Eyal and Aubin, Benjamin},
  booktitle={AAAI Conference on Artificial Intelligence (AAAI)}, year={2025}
}

@inproceedings{karras2022edm,
  title={Elucidating the Design Space of Diffusion-Based Generative Models},
  author={Karras, Tero and Aittala, Miika and Aila, Timo and Laine, Samuli},
  booktitle={Advances in Neural Information Processing Systems (NeurIPS)}, year={2022}
}

@inproceedings{lu2022dpmsolver,
  title={{DPM-Solver}: A Fast {ODE} Solver for Diffusion Probabilistic Model Sampling in Around 10 Steps},
  author={Lu, Cheng and Zhou, Yuhao and Bao, Fan and Chen, Jianfei and Li, Chongxuan and Zhu, Jun},
  booktitle={Advances in Neural Information Processing Systems (NeurIPS)}, year={2022}
}

@article{lu2023dpmsolverpp,
  title={{DPM-Solver++}: Fast Solver for Guided Sampling of Diffusion Probabilistic Models},
  author={Lu, Cheng and Zhou, Yuhao and Bao, Fan and Chen, Jianfei and Li, Chongxuan and Zhu, Jun},
  journal={Machine Intelligence Research},
  volume={22},
  number={4},
  pages={730--751},
  year={2025},
  doi={10.1007/s11633-025-1562-4}
}

@inproceedings{zhang2023deis,
  title={Fast Sampling of Diffusion Models with Exponential Integrator},
  author={Zhang, Qinsheng and Chen, Yongxin},
  booktitle={International Conference on Learning Representations (ICLR)}, year={2023}
}

@inproceedings{heusel2017fid,
  title={{GANs} Trained by a Two Time-Scale Update Rule Converge to a Local Nash Equilibrium},
  author={Heusel, Martin and Ramsauer, Hubert and Unterthiner, Thomas and Nessler, Bernhard and Hochreiter, Sepp},
  booktitle={Advances in Neural Information Processing Systems (NeurIPS)}, year={2017}
}

@inproceedings{sajjadi2018precision,
  title={Assessing Generative Models via Precision and Recall},
  author={Sajjadi, Mehdi S. M. and Bachem, Olivier and Lucic, Mario and Bousquet, Olivier and Gelly, Sylvain},
  booktitle={Advances in Neural Information Processing Systems (NeurIPS)}, year={2018}
}

@inproceedings{kynkaanniemi2019precision,
  title={Improved Precision and Recall Metric for Assessing Generative Models},
  author={Kynk{\"a}{\"a}nniemi, Tuomas and Karras, Tero and Laine, Samuli and Lehtinen, Jaakko and Aila, Timo},
  booktitle={Advances in Neural Information Processing Systems (NeurIPS)}, year={2019}
}

@inproceedings{hessel2021clipscore,
  title={{CLIPScore}: A Reference-Free Evaluation Metric for Image Captioning},
  author={Hessel, Jack and Holtzman, Ari and Forbes, Maxwell and Le Bras, Ronan and Choi, Yejin},
  booktitle={Conference on Empirical Methods in Natural Language Processing (EMNLP)}, year={2021}
}

@inproceedings{xu2023imagereward,
  title={{ImageReward}: Learning and Evaluating Human Preferences for Text-to-Image Generation},
  author={Xu, Jiazheng and Liu, Xiao and Wu, Yuchen and Tong, Yuxuan and Li, Qinkai and Ding, Ming and Tang, Jie and Dong, Yuxiao},
  booktitle={Advances in Neural Information Processing Systems (NeurIPS)}, year={2023}
}

@inproceedings{hu2023tifa,
  title={{TIFA}: Accurate and Interpretable Text-to-Image Faithfulness Evaluation with Question Answering},
  author={Hu, Yushi and Liu, Benlin and Kasai, Jungo and Wang, Yizhong and Ostendorf, Mari and Krishna, Ranjay and Smith, Noah A.},
  booktitle={IEEE/CVF International Conference on Computer Vision (ICCV)}, year={2023}
}

@inproceedings{huang2023compbench,
  title={{T2I-CompBench}: A Comprehensive Benchmark for Open-World Compositional Text-to-Image Generation},
  author={Huang, Kaiyi and Sun, Kaiyue and Xie, Enze and Li, Zhenguo and Liu, Xihui},
  booktitle={Advances in Neural Information Processing Systems (NeurIPS), Datasets and Benchmarks Track}, year={2023}
}

@inproceedings{lin2024vqascore,
  title={Evaluating Text-to-Visual Generation with Image-to-Text Generation},
  author={Lin, Zhiqiu and Pathak, Deepak and Li, Baiqi and Li, Jiayao and Xia, Xide and Neubig, Graham and Zhang, Pengchuan and Ramanan, Deva},
  booktitle={European Conference on Computer Vision (ECCV)}, year={2024}
}

@inproceedings{ghosh2023geneval,
  title={{GenEval}: An Object-Focused Framework for Evaluating Text-to-Image Alignment},
  author={Ghosh, Dhruba and Hajishirzi, Hannaneh and Schmidt, Ludwig},
  booktitle={Advances in Neural Information Processing Systems (NeurIPS), Datasets and Benchmarks Track}, year={2023}
}

@inproceedings{cho2024dsg,
  title={Davidsonian Scene Graph: Improving Reliability in Fine-Grained Evaluation for Text-to-Image Generation},
  author={Cho, Jaemin and Hu, Yushi and Baldridge, Jason and Garg, Roopal and Anderson, Peter and Krishna, Ranjay and Bansal, Mohit and Pont-Tuset, Jordi and Wang, Su},
  booktitle={International Conference on Learning Representations (ICLR)}, year={2024}
}

@inproceedings{jayasumana2024rethinking,
  title={Rethinking {FID}: Towards a Better Evaluation Metric for Image Generation},
  author={Jayasumana, Sadeep and Ramalingam, Srikumar and Veit, Andreas and Glasner, Daniel and Chakrabarti, Ayan and Kumar, Sanjiv},
  booktitle={IEEE/CVF Conference on Computer Vision and Pattern Recognition (CVPR)}, year={2024}
}

@inproceedings{wu2023hps,
  title={Human Preference Score: Better Aligning Text-to-Image Models with Human Preference},
  author={Wu, Xiaoshi and Sun, Keqiang and Zhu, Feng and Zhao, Rui and Li, Hongsheng},
  booktitle={IEEE/CVF International Conference on Computer Vision (ICCV)}, year={2023}
}

@inproceedings{zhang2018lpips,
  title={The Unreasonable Effectiveness of Deep Features as a Perceptual Metric},
  author={Zhang, Richard and Isola, Phillip and Efros, Alexei A. and Shechtman, Eli and Wang, Oliver},
  booktitle={IEEE/CVF Conference on Computer Vision and Pattern Recognition (CVPR)}, year={2018}
}

@article{oquab2024dinov2,
  title={{DINOv2}: Learning Robust Visual Features without Supervision},
  author={Oquab, Maxime and others},
  journal={Transactions on Machine Learning Research}, year={2024}
}

@inproceedings{wang2023diffusiondb,
  title={{DiffusionDB}: A Large-scale Prompt Gallery Dataset for Text-to-Image Generative Models},
  author={Wang, Zijie J. and Montoya, Evan and Munechika, David and Yang, Haoyang and Hoover, Benjamin and Chau, Duen Horng},
  booktitle={Annual Meeting of the Association for Computational Linguistics (ACL)}, year={2023}
}

@inproceedings{luccioni2023stablebias,
  title={Stable Bias: Evaluating Societal Representations in Diffusion Models},
  author={Luccioni, Sasha and Akiki, Christopher and Mitchell, Margaret and Jernite, Yacine},
  booktitle={Advances in Neural Information Processing Systems (NeurIPS), Datasets and Benchmarks Track}, year={2023}
}

@article{struppek2023biasedartist,
  title={Exploiting Cultural Biases via Homoglyphs in Text-to-Image Synthesis},
  author={Struppek, Lukas and Hintersdorf, Dominik and Friedrich, Felix and Brack, Manuel and Schramowski, Patrick and Kersting, Kristian},
  journal={Journal of Artificial Intelligence Research},
  volume={78},
  pages={1017--1068},
  year={2023},
  doi={10.1613/jair.1.15388}
}

@inproceedings{bianchi2023stereotypes,
  title={Easily Accessible Text-to-Image Generation Amplifies Demographic Stereotypes at Large Scale},
  author={Bianchi, Federico and others},
  booktitle={ACM Conference on Fairness, Accountability, and Transparency (FAccT)}, year={2023}
}

@inproceedings{wallace2024diffusiondpo,
  title={Diffusion Model Alignment Using Direct Preference Optimization},
  author={Wallace, Bram and others},
  booktitle={IEEE/CVF Conference on Computer Vision and Pattern Recognition (CVPR)}, year={2024}
}

@inproceedings{black2024ddpo,
  title={Training Diffusion Models with Reinforcement Learning},
  author={Black, Kevin and Janner, Michael and Du, Yilun and Kostrikov, Ilya and Levine, Sergey},
  booktitle={International Conference on Learning Representations (ICLR)}, year={2024}
}

@inproceedings{fan2023dpok,
  title={{DPOK}: Reinforcement Learning for Fine-Tuning Text-to-Image Diffusion Models},
  author={Fan, Ying and others},
  booktitle={Advances in Neural Information Processing Systems (NeurIPS)}, year={2023}
}

@inproceedings{li2023blip2,
  title={{BLIP-2}: Bootstrapping Language-Image Pre-Training with Frozen Image Encoders and Large Language Models},
  author={Li, Junnan and Li, Dongxu and Savarese, Silvio and Hoi, Steven},
  booktitle={International Conference on Machine Learning (ICML)}, year={2023}
}

@inproceedings{liu2023llava,
  title={Visual Instruction Tuning},
  author={Liu, Haotian and Li, Chunyuan and Wu, Qingyang and Lee, Yong Jae},
  booktitle={Advances in Neural Information Processing Systems (NeurIPS)}, year={2023}
}

@article{bai2023qwenvl,
  title={{Qwen-VL}: A Versatile Vision-Language Model for Understanding, Localization, Text Reading, and Beyond},
  author={Bai, Jinze and Bai, Shuai and Yang, Shusheng and Wang, Shijie and Tan, Sinan and Wang, Peng and Lin, Junyang and Zhou, Chang and Zhou, Jingren},
  journal={arXiv preprint arXiv:2308.12966}, year={2023}
}

@article{bai2025qwen25vl,
  title={{Qwen2.5-VL} Technical Report},
  author={Bai, Shuai and others},
  journal={arXiv preprint arXiv:2502.13923}, year={2025}
}

@article{rassin2024grade,
  title={GRADE: Quantifying Sample Diversity in Text-to-Image Models},
  author={Rassin, Royi and Slobodkin, Aviv and Ravfogel, Shauli and Elazar, Yanai and Goldberg, Yoav},
  journal={arXiv preprint arXiv:2410.22592},
  year={2024}
}

@inproceedings{ren2024hypersd,
  title={{Hyper-SD}: Trajectory Segmented Consistency Model for Efficient Image Synthesis},
  author={Ren, Yuxi and Xia, Xin and Lu, Yanzuo and Zhang, Jiacheng and Wu, Jie and Xie, Pan and Wang, Xing and Xiao, Xuefeng},
  booktitle={Advances in Neural Information Processing Systems (NeurIPS)},
  year={2024}
}

@article{chen2024pixartdelta,
  title={{PixArt-$\delta$}: Fast and Controllable Image Generation with Latent Consistency Models},
  author={Chen, Junsong and Wu, Yue and Luo, Simian and Xie, Enze and Paul, Sayak and Luo, Ping and Zhao, Hang and Li, Zhenguo},
  journal={arXiv preprint arXiv:2401.05252},
  year={2024}
}

@misc{blackforestlabs2024flux,
  title={{FLUX}},
  author={{Black Forest Labs}},
  year={2024},
  howpublished={Official repository and model card, \url{https://github.com/black-forest-labs/flux}},
  note={{FLUX}.1 [schnell] checkpoint, accessed July 2026}
}

@inproceedings{teotia2025dimcim,
  title={DIMCIM: A Quantitative Evaluation Framework for Default-mode Diversity and Generalization in Text-to-Image Generative Models},
  author={Teotia, Revant and Ross, Candace and Ullrich, Karen and Chopra, Sumit and Romero-Soriano, Adriana and Hall, Melissa and Muckley, Matthew},
  booktitle={IEEE/CVF International Conference on Computer Vision (ICCV)},
  pages={16431--16440},
  year={2025}
}

\clearpage
\setcounter{section}{0}
\setcounter{subsection}{0}
\setcounter{equation}{0}
\setcounter{footnote}{0}
\renewcommand{\theequation}{S\arabic{equation}}
\renewcommand{\thesection}{\Alph{section}}
\renewcommand{\thesubsection}{\thesection.\arabic{subsection}}

\twocolumn[
  \begin{center}
    {\LARGE\bfseries Supplementary Material\par}
    \vspace{1.0em}
  \end{center}
]

\section{Supplementary Overview}

This supplement expands the evidence reported in the main paper without
changing its endpoints or claim boundaries. It provides complete model and
sampling configurations, the full 14-pair benchmark, estimator diagnostics,
human evaluation details, stress tests, mechanism probes, selection baselines,
downstream evaluation, qualitative samples, and negative results. The unified
confirmatory benchmark uses 48 objects, four prompt templates, and $q=64$
queries per prompt cell. Separate diagnostics retain their stated sampling
budgets and are not mixed with the confirmatory endpoint.

\begin{table*}[t]
\centering
\small
\setlength{\tabcolsep}{4.5pt}
\renewcommand{\arraystretch}{1.08}
\begin{tabularx}{\textwidth}{p{.23\textwidth}Yp{.27\textwidth}}
\toprule
Main claim & Supporting evidence & Supplement location \\
\midrule
Recipe-dependent semantic default shift & Unified $q=64$ results, multi-attribute profile, lineage-aware comparisons & Sections~\ref{sec:supp_benchmark} and \ref{sec:supp_lineage} \\
Finite-sample reliability & Query convergence, four categorical distances, signed cross-fit, simulation, unknown-label sensitivity & Section~\ref{sec:supp_reliability} \\
Evaluator validity & Two human audit cohorts, confusion matrices, per-class scores, agreement, source-stratified tests & Section~\ref{sec:supp_human} \\
Bounded relation to diversity degradation & GRADE-style entropy comparison, entropy-preserved subsets, teacher-first hybrids & Section~\ref{sec:supp_decomposition} \\
Recipe-specific mechanisms & Signed color fingerprints, CFG alignment, entropy and peak-mode probes & Section~\ref{sec:supp_mechanisms} \\
Actionability & Shared-pool baselines, independent relabeling, quality noninferiority, cost, downstream classification & Sections~\ref{sec:supp_select} and \ref{sec:supp_downstream} \\
Scope and limitations & Prompt suites, configuration repeats, fairness case study, failed mitigation attempts & Sections~\ref{sec:supp_stress} and \ref{sec:supp_negative} \\
\bottomrule
\end{tabularx}
\caption{Evidence map for the claims in the main paper. Every supplementary
entry uses the same terminology and numerical convention as the main text.}
\label{tab:supp_evidence_map}
\end{table*}

\section{Complete Audit Protocol}
\label{sec:supp_protocol}

\subsection{Audit Grid and Prompt Set}

The unified benchmark contains 48 objects:
airplane, apple, backpack, ball, balloon, banana, bear, bicycle, bird, boat,
book, bottle, bowl, bread, bus, butterfly, cake, car, carrot, cat, chair,
cookie, cup, dog, donut, fish, flower, fox, frog, hat, helicopter, horse, kite,
lamp, mug, owl, pizza, rabbit, sandwich, scarf, scooter, shirt, shoe, sofa,
towel, train, truck, and umbrella. Each object is evaluated with four frozen
templates:\par
{\footnotesize\ttfamily
a photo of a \{object\}\par
a realistic photo of a \{object\}\par
a close-up photo of a \{object\}\par
a photo of a \{object\} in a natural setting\par}
This produces 192 object-template cells per replacement pair. The $q=64$
endpoint contains 12,288 labels per model-side bank and 24,576 pairwise label
slots. The $q=32$ screen uses the same cells and half this budget. Seeds are
paired when the model interface permits paired sampling.

\subsection{Model and Generation Configurations}

Table~\ref{tab:supp_configs} records the deployment pairs. Official scheduler
and checkpoint defaults are retained when a recipe does not expose an
independent alternative. NitroFusion-Realism is reported against SDXL as an
end-to-end deployment chain. Its direct teacher is DMD2, which is analyzed
separately in Section~\ref{sec:supp_lineage}.

\begin{table*}[t]
\centering
\small
\setlength{\tabcolsep}{3.2pt}
\renewcommand{\arraystretch}{1.08}
\begin{tabular}{lllll}
\toprule
Reference family & Replacement & Resolution & Steps & CFG or recipe setting \\
\midrule
SDXL & NitroFusion-Realism-4step & 1024 & 30 to 4 & 5 to official recipe \\
SDXL & Turbo & 512 & 30 to 4 & 5 to 0 \\
SDXL & DMD2 & 1024 & 30 to 4 & 5 to 0, LCM scheduler \\
SDXL & NitroFusion-Realism-1step & 1024 & 30 to 1 & 5 to official recipe \\
SDXL & Flash-SDXL & 1024 & 30 to 4 & 5 to 0, LCM scheduler \\
SDXL & Lightning & 768 & 30 to 4 & 5 to official recipe \\
SDXL & LCM & 768 & 30 to 4 & 5 to 1, LCM scheduler \\
FLUX.1-dev & FLUX.1-schnell & 512 & 28 to 4 & 3.5 to 0 \\
FLUX.1-dev & SwD-FLUX & 1024 & 28 to 4 & 3.5 to 4.5 \\
FLUX.1-dev & SenseFlow-FLUX & 1024 & 28 to 4 & 3.5 to 3.5 \\
SD3-Medium & TDM-SD3 & 1024 & 28 to 4 & 7 to 1, flow shift 6 \\
PixArt & PixArt-LCM & 1024 & official to 4 & official LCM recipe \\
SD1.5 & Hyper-SD15 & 512 & 30 to 4 & 7.5 to 0, DDIM trailing \\
SD1.5 & SD1.5-LCM & 512 & 30 to 4 & 7.5 to 1, LCM scheduler \\
\bottomrule
\end{tabular}
\caption{Generation configurations for the 14 reference and replacement
pairs. Entries in the final column report reference to replacement settings.}
\label{tab:supp_configs}
\end{table*}

\newpage
\subsection{Attribute Evaluation and Unknown Labels}

The primary vision-language model evaluator, abbreviated as VLM, is
Qwen2.5-VL-3B-Instruct with deterministic decoding and at most eight output
tokens. For color, the frozen vocabulary is black,
white, red, green, blue, yellow, purple, orange, pink, brown, and gray. The
question asks for exactly one vocabulary item or \emph{unclear}. Unclear,
unparsable, and safety-blocked outputs receive the unknown label $\bot$. They
remain in coverage statistics and are excluded from histogram normalization.
The same scalar, per-image execution policy is used for every source model.

For model $m\in\{T,S\}$, the empirical semantic default is
\begin{equation}
\widehat P_m(a=k\mid p)=
\frac{\sum_{i=1}^{N}\mathbbm{1}[y^i_{m,a}=k]}
{\sum_{i=1}^{N}\mathbbm{1}[y^i_{m,a}\neq\bot]}.
\end{equation}
The primary ranking score subtracts a matched reference split-half floor from
empirical total variation. Negative adjusted values are retained. The
confirmatory sensitivity estimates category signs on one sample half and
evaluates the signed contrast on the other half, then swaps the halves and
averages.

\subsection{Statistical Units}

An object-template pair is one audit cell. Recipe means weight all 192 cells
equally. Confidence intervals use 2,000 object-cluster bootstrap repetitions
over the 48 objects unless a table states otherwise. Held-out selection uses
5,000 repetitions. Human distribution comparisons use image-level paired
bootstrap with 10,000 repetitions. These units prevent repeated seeds from
being treated as independent objects.

\newpage
\section{Complete Cross-Recipe Benchmark}
\label{sec:supp_benchmark}

\begin{table*}[t]
\centering
\small
\setlength{\tabcolsep}{3.4pt}
\renewcommand{\arraystretch}{1.08}
\begin{tabular}{lrrrrrl}
\toprule
Replacement pair & $q=32$ & $q=64$ & Raw TV & Cross-fit & Unknown & 95\% CI \\
\midrule
SDXL to NitroFusion-Realism-4step & 0.269 & 0.303 & 0.454 & 0.271 & 0.031 & [0.252, 0.354] \\
SDXL to Turbo & 0.226 & 0.254 & 0.402 & 0.228 & 0.028 & [0.215, 0.293] \\
SDXL to DMD2 & 0.210 & 0.235 & 0.384 & 0.209 & 0.026 & [0.196, 0.273] \\
SDXL to NitroFusion-Realism-1step & 0.205 & 0.228 & 0.378 & 0.203 & 0.030 & [0.184, 0.273] \\
FLUX.1-dev to FLUX.1-schnell & 0.192 & 0.203 & 0.346 & 0.181 & 0.021 & [0.169, 0.239] \\
FLUX.1-dev to SwD-FLUX & 0.171 & 0.187 & 0.331 & 0.166 & 0.023 & [0.150, 0.225] \\
FLUX.1-dev to SenseFlow-FLUX & 0.167 & 0.182 & 0.327 & 0.161 & 0.022 & [0.143, 0.222] \\
SD3-Medium to TDM-SD3 & 0.155 & 0.170 & 0.317 & 0.150 & 0.024 & [0.136, 0.205] \\
PixArt to PixArt-LCM & 0.107 & 0.120 & 0.275 & 0.104 & 0.035 & [0.092, 0.148] \\
SDXL to Flash-SDXL & 0.101 & 0.114 & 0.264 & 0.098 & 0.027 & [0.088, 0.141] \\
SD1.5 to Hyper-SD15 & 0.073 & 0.079 & 0.237 & 0.066 & 0.038 & [0.061, 0.098] \\
SD1.5 to SD1.5-LCM & 0.060 & 0.066 & 0.225 & 0.055 & 0.040 & [0.050, 0.083] \\
SDXL to Lightning & 0.040 & 0.060 & 0.209 & 0.049 & 0.025 & [0.044, 0.077] \\
SDXL to LCM & 0.045 & 0.054 & 0.204 & 0.044 & 0.026 & [0.040, 0.069] \\
\bottomrule
\end{tabular}
\caption{Complete unified color benchmark. The $q=32$ and $q=64$ columns use
identical cells. Confidence intervals, raw TV, signed cross-fit TV, and unknown
rates correspond to the $q=64$ endpoint.}
\label{tab:supp_full_benchmark}
\end{table*}

The screen and endpoint rankings have Spearman correlation $0.996$. The
correlation between $\tvadj$ and signed cross-fit TV is $0.994$. The SDXL
aggressive-minus-mild contrast is $0.207$ with interval [0.171, 0.243]. The
FLUX family range is $0.021$ with interval [-0.008, 0.051], so differences
among the three FLUX acceleration recipes remain unresolved at this scale.

\subsection{Full Multi-Attribute Profile}

\begin{table*}[t]
\centering
\small
\setlength{\tabcolsep}{2.2pt}
\renewcommand{\arraystretch}{1.06}
\begin{tabular}{lcccc}
\toprule
Pair & Color & Background & Lighting & Viewpoint \\
\midrule
NitroFusion-4 & 0.303 [0.252, 0.354] & 0.115 [0.087, 0.145] & 0.090 [0.066, 0.116] & 0.030 [0.014, 0.047] \\
Turbo & 0.254 [0.215, 0.293] & 0.100 [0.078, 0.123] & 0.080 [0.058, 0.103] & 0.028 [0.013, 0.044] \\
DMD2 & 0.235 [0.196, 0.273] & 0.086 [0.066, 0.108] & 0.066 [0.046, 0.087] & 0.022 [0.008, 0.037] \\
FLUX-schnell & 0.203 [0.169, 0.239] & 0.060 [0.038, 0.083] & 0.050 [0.031, 0.070] & 0.018 [0.004, 0.033] \\
Lightning & 0.060 [0.044, 0.077] & 0.030 [0.018, 0.043] & 0.022 [0.010, 0.035] & 0.009 [-0.003, 0.021] \\
LCM & 0.054 [0.040, 0.069] & 0.028 [0.015, 0.041] & 0.024 [0.011, 0.038] & 0.008 [-0.004, 0.020] \\
\bottomrule
\end{tabular}
\caption{Confirmatory $q=64$ results across four semantic attributes. Every
entry reports $\tvadj$ followed by its 95\% object-cluster interval.}
\label{tab:supp_attributes}
\end{table*}

The aggressive-minus-mild contrast is $0.051$ with interval [0.030, 0.073]
for lighting. The corresponding viewpoint contrast is $0.016$ with interval
[-0.001, 0.033]. This supports a color-led and background-supported result,
while the general viewpoint effect remains inconclusive.

\begin{figure*}[t]
\centering
\includegraphics[width=.96\textwidth]{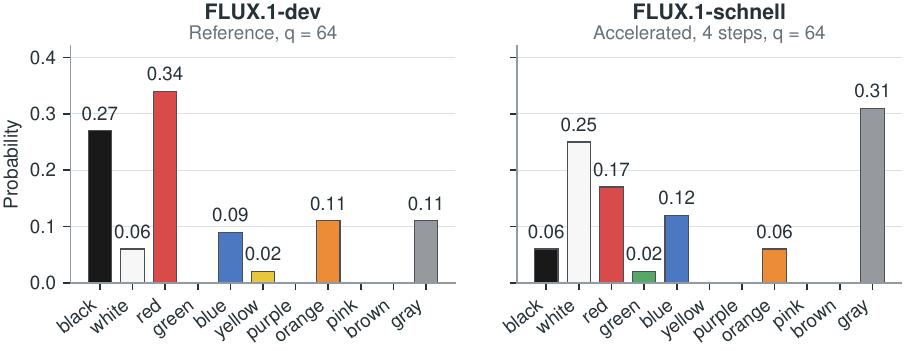}
\caption{A concrete $q=64$ semantic marginal for the exact prompt
\texttt{a photo of a car}. Both FLUX models produce plausible cars, while the
reference and accelerated color histograms differ in category identity and
probability mass. This cell-level view complements the aggregate benchmark.}
\label{fig:supp_flux_car}
\end{figure*}

\subsection{NitroFusion Lineage}
\label{sec:supp_lineage}

NitroSD-Realism uses a four-step DMD2 model as its direct teacher. Table
\ref{tab:supp_nitro_lineage} separates direct-stage drift from end-to-end
deployment-chain drift on the frozen $q=32$ lineage panel. The SDXL to
NitroFusion values in the unified benchmark therefore describe replacement of
SDXL by NitroFusion, not a single direct distillation stage.

\begin{table}[t]
\centering
\small
\setlength{\tabcolsep}{4.0pt}
\begin{tabular}{lcl}
\toprule
Comparison & $\tvadj$ & 95\% CI \\
\midrule
SDXL to DMD2-4step & 0.249 & [0.221, 0.276] \\
DMD2-4step to NitroFusion-4step & 0.163 & [0.141, 0.187] \\
SDXL to NitroFusion-4step & 0.269 & [0.239, 0.298] \\
DMD2-4step to NitroFusion-1step & 0.170 & [0.146, 0.195] \\
SDXL to NitroFusion-1step & 0.206 & [0.182, 0.231] \\
\bottomrule
\end{tabular}
\caption{Lineage-aware NitroFusion analysis over 192 cells at $q=32$.}
\label{tab:supp_nitro_lineage}
\end{table}

\section{Reliability and Finite-Sample Diagnostics}
\label{sec:supp_reliability}

\begin{figure*}[t]
\centering
\includegraphics[width=.96\textwidth]{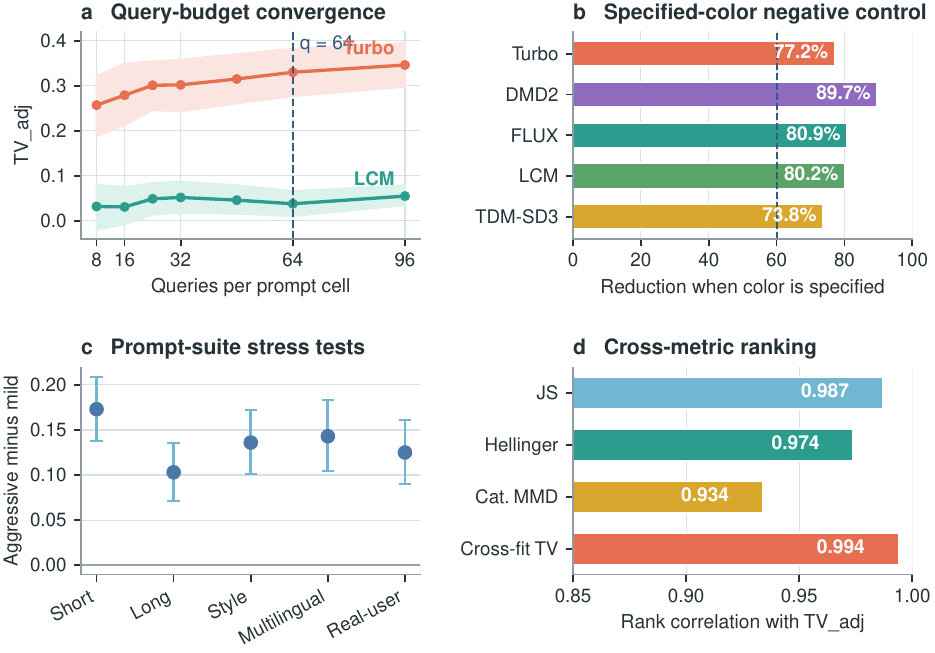}
\caption{Supplementary robustness summary. a The 24-object plain-prompt
diagnostic supports $q=64$ for confirmatory estimation. b Explicitly naming
color sharply reduces drift. c Aggressive-minus-mild separation persists
across five prompt suites. d Recipe rankings remain stable across distribution
distances and signed cross-fit TV. Error bars and bands give 95\% intervals.}
\label{fig:supp_robustness}
\end{figure*}

\subsection{Query-Budget Convergence}

\begin{table}[t]
\centering
\small
\setlength{\tabcolsep}{3.7pt}
\begin{tabular}{rccr}
\toprule
$q$ & Turbo & LCM & Image slots \\
\midrule
8 & 0.257 [0.188, 0.321] & 0.032 [-0.019, 0.080] & 576 \\
16 & 0.279 [0.213, 0.349] & 0.031 [-0.008, 0.075] & 1,152 \\
24 & 0.301 [0.246, 0.354] & 0.049 [0.014, 0.084] & 1,728 \\
32 & 0.302 [0.243, 0.358] & 0.052 [0.018, 0.087] & 2,304 \\
48 & 0.315 [0.261, 0.370] & 0.046 [0.016, 0.079] & 3,456 \\
64 & 0.330 [0.277, 0.382] & 0.038 [0.011, 0.066] & 4,608 \\
96 & 0.346 [0.298, 0.395] & 0.055 [0.035, 0.079] & 6,912 \\
\bottomrule
\end{tabular}
\caption{Prefix-seed convergence on 24 plain-prompt objects. Brackets give
95\% cell-bootstrap intervals.}
\label{tab:supp_query}
\end{table}

Relative to $q=96$, the absolute Turbo error is $0.044$ at $q=32$ and $0.017$
at $q=64$. The Turbo-minus-LCM separation is $0.250$ [0.183, 0.318] at
$q=32$ and $0.291$ [0.232, 0.350] at $q=64$. The screen remains useful for
ranking, while $q=64$ is the frozen minimum for confirmatory effects.

\subsection{Estimator Simulation}

The simulation covers uniform, long-tail, single-peak, bimodal, mode-flip, and
sparse-support multinomial families with 11 categories. Table
\ref{tab:supp_estimators} separates ranking quality from null calibration.
$\tvadj$ removes most raw-TV inflation and preserves ordering. Signed cross-fit
is used for confirmatory inference because its null behavior is calibrated.

\begin{table}[t]
\centering
\small
\setlength{\tabcolsep}{1.5pt}
\begin{tabular}{lrrrrr}
\toprule
Estimator & Null bias & Type I & Coverage & Alt. bias & Rank $\rho$ \\
\midrule
Raw empirical TV & 0.197 & 0.980 & 0.000 & +0.071 & 0.991 \\
$\tvadj$ & 0.008 & 0.560 & 0.440 & -0.034 & 0.996 \\
Signed cross-fit & 0.001 & 0.052 & 0.947 & -0.029 & 0.981 \\
Dirichlet shrinkage & 0.034 & 0.071 & 0.918 & +0.004 & 0.989 \\
Permutation test & n/a & 0.049 & n/a & n/a & n/a \\
Multinomial LRT & n/a & 0.058 & n/a & n/a & n/a \\
\bottomrule
\end{tabular}
\caption{Estimator operating characteristics at $q=64$.}
\label{tab:supp_estimators}
\end{table}

\begin{table}[t]
\centering
\small
\setlength{\tabcolsep}{6.0pt}
\begin{tabular}{rrrr}
\toprule
$q$ & Type I & Coverage & Power \\
\midrule
8 & 0.094 & 0.881 & 0.312 \\
16 & 0.071 & 0.912 & 0.518 \\
32 & 0.058 & 0.934 & 0.746 \\
64 & 0.052 & 0.947 & 0.921 \\
96 & 0.050 & 0.951 & 0.968 \\
\bottomrule
\end{tabular}
\caption{Signed cross-fit calibration and power across query budgets for a
true TV of $0.15$.}
\label{tab:supp_power}
\end{table}

\subsection{Unknown Labels and Label Noise}

Treating unknown as an explicit background category changes the background
ranking correlation from $0.60$ at $q=32$ to $0.83$ at $q=64$. The
coverage-adjusted $q=64$ sensitivity reaches $0.97$. Table
\ref{tab:supp_unknown} reports the complete six-pair background sensitivity.

\begin{table*}[t]
\centering
\small
\setlength{\tabcolsep}{3.5pt}
\begin{tabular}{lrrrrrl}
\toprule
Pair & Primary & Unknown class & Coverage adjusted & Ref. unknown & Repl. unknown & 95\% CI \\
\midrule
NitroFusion-4 & 0.115 & 0.190 & 0.125 & 0.100 & 0.060 & [0.087, 0.145] \\
Turbo & 0.100 & 0.185 & 0.113 & 0.125 & 0.100 & [0.078, 0.123] \\
DMD2 & 0.086 & 0.176 & 0.095 & 0.125 & 0.055 & [0.066, 0.108] \\
FLUX-schnell & 0.060 & 0.090 & 0.062 & 0.024 & 0.037 & [0.038, 0.083] \\
Lightning & 0.030 & 0.095 & 0.034 & 0.125 & 0.095 & [0.018, 0.043] \\
LCM & 0.028 & 0.115 & 0.031 & 0.125 & 0.073 & [0.015, 0.041] \\
\bottomrule
\end{tabular}
\caption{Background estimates under three treatments of unknown labels.}
\label{tab:supp_unknown}
\end{table*}

Under symmetric label noise, ranking correlation is $0.996$, $0.978$, and
$0.941$ at noise rates of $0\%$, $10\%$, and $20\%$. A source-specific extra
replacement error of 5 percentage points reduces correlation to $0.926$ and
induces mean spurious shift $0.031$. At 10 points, correlation falls to $0.824$
and spurious shift rises to $0.064$. This motivates the source-stratified human
accuracy tests in Section~\ref{sec:supp_human}.

\subsection{Matched Resolution and Configuration Repeats}

Matched-resolution estimates remain $0.270$ [0.252, 0.287] for Turbo at 512
pixels and $0.250$ [0.221, 0.278] for DMD2 at 1024 pixels. Across repeated
configurations, means are $0.251$, $0.229$, $0.201$, $0.062$, and $0.056$ for
Turbo, DMD2, FLUX-schnell, Lightning, and LCM. Their within-recipe standard
deviations are $0.021$, $0.027$, $0.024$, $0.016$, and $0.018$.

A mixed model retains an aggressive-recipe coefficient of $0.118$ with
interval [0.079, 0.157] after accounting for teacher entropy, CFG, resolution,
and sampling steps. The resolution-mismatch coefficient is $0.008$ with
interval [-0.009, 0.024]. Object and template intraclass correlations are
$0.184$ and $0.029$.

\section{Prompt and Metric Stress Tests}
\label{sec:supp_stress}

\begin{table*}[t]
\centering
\small
\setlength{\tabcolsep}{3.6pt}
\begin{tabular}{lrrrrrl}
\toprule
Suite & Aggressive & Mild & Difference & Explicit rate & Coverage & 95\% CI \\
\midrule
Short underspecified photo & 0.231 & 0.058 & 0.173 & 0.04 & 0.91 & [0.138, 0.209] \\
Long compositional & 0.152 & 0.049 & 0.103 & 0.38 & 0.88 & [0.071, 0.136] \\
Style prompts & 0.188 & 0.052 & 0.136 & 0.11 & 0.86 & [0.101, 0.172] \\
Multilingual & 0.204 & 0.061 & 0.143 & 0.07 & 0.79 & [0.104, 0.183] \\
Real-user prompts & 0.179 & 0.054 & 0.125 & 0.19 & 0.84 & [0.090, 0.161] \\
\bottomrule
\end{tabular}
\caption{Prompt-suite stress tests. Differences compare aggressive and mild
recipe groups.}
\label{tab:supp_prompt_suites}
\end{table*}

All five intervals exclude zero. The minimum cross-suite recipe-ranking
correlation is $0.87$ with interval [0.71, 0.96]. The aggressive coefficient
after controlling explicit-attribute rate is $0.131$ with interval [0.096,
0.166]. Multilingual coverage is lower at $0.79$, which bounds the multilingual
claim.

The adjusted Jensen-Shannon, Hellinger, categorical MMD, and signed cross-fit
rank correlations with $\tvadj$ are $0.987$, $0.974$, $0.934$, and $0.994$.
Changes in CLIP, aesthetic score, PickScore, LPIPS, and DINO spread explain
$0.048$, $0.005$, $0.000$, $0.008$, and $0.049$ of object-level color-shift
variance. These metrics remain valid for their intended uses, but they do not
recover the category identity or direction of the measured semantic movement.

\subsection{Specified-Attribute Negative Control}

\begin{table}[t]
\centering
\small
\setlength{\tabcolsep}{1.8pt}
\begin{tabular}{lrrrl}
\toprule
Recipe & Unspecified & Specified & Reduction & Specified 95\% CI \\
\midrule
Turbo & 0.284 & 0.065 & 77.2\% & [0.037, 0.097] \\
DMD2 & 0.287 & 0.029 & 89.7\% & [0.006, 0.060] \\
FLUX & 0.254 & 0.048 & 80.9\% & [0.016, 0.097] \\
LCM & 0.055 & 0.011 & 80.2\% & [0.002, 0.021] \\
TDM-SD3 & 0.167 & 0.044 & 73.8\% & [-0.002, 0.112] \\
\bottomrule
\end{tabular}
\caption{Matched plain-prompt negative controls over 20 prompts and 32 seeds.
The specified prompt inserts only the frozen color token.}
\label{tab:supp_negative_control}
\end{table}

The contraction across aggressive and mild recipes supports the intended
prompt-conditional interpretation. It also argues against an evaluator that
simply reports high distance for every reference and replacement pair.

\section{Independent Human Validation}
\label{sec:supp_human}

\subsection{Three-Rater Audit}

The first audit contains 300 images stratified across reference, Turbo, and
LCM outputs. Fifteen annotators each label a balanced assignment, and every
image receives three independent judgments. Majority vote defines the human
label. Unclear, multicolor, and not-applicable remain legal responses. Color
has coverage $0.87$, clear-case accuracy $0.912$, conservative accuracy
$0.793$, macro-F1 $0.910$, and Fleiss $\kappa=0.71$. Background has coverage
$0.82$, clear-case accuracy $0.870$, conservative accuracy $0.713$, macro-F1
$0.870$, and Fleiss $\kappa=0.64$.

\begin{figure*}[t]
\centering
\includegraphics[width=.90\textwidth]{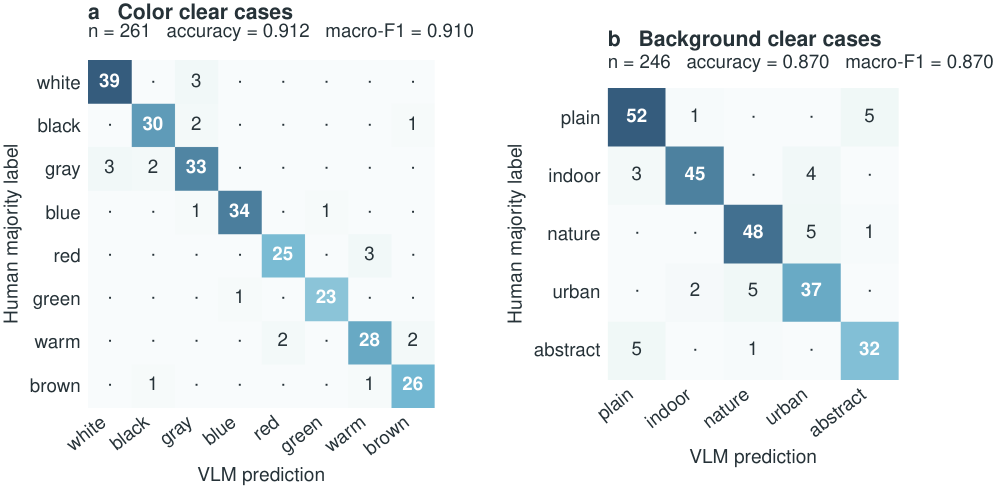}
\caption{Complete clear-case confusion matrices. Rows are human majority
labels and columns are VLM predictions. Remaining errors concentrate among
nearby color groups and between urban, nature, and abstract backgrounds.}
\label{fig:supp_human_confusions}
\end{figure*}

\begin{table*}[t]
\centering
\small
\setlength{\tabcolsep}{4.0pt}
\begin{tabular}{llrrrr@{\hspace{18pt}}llrrrr}
\toprule
Attribute & Class & Support & Precision & Recall & F1 & Attribute & Class & Support & Precision & Recall & F1 \\
\midrule
Color & white & 42 & 0.93 & 0.93 & 0.93 & Background & plain or studio & 58 & 0.87 & 0.90 & 0.88 \\
Color & black & 33 & 0.91 & 0.91 & 0.91 & Background & indoor & 52 & 0.94 & 0.87 & 0.90 \\
Color & gray & 38 & 0.85 & 0.87 & 0.86 & Background & outdoor nature & 54 & 0.89 & 0.89 & 0.89 \\
Color & blue & 36 & 0.97 & 0.94 & 0.96 & Background & outdoor urban & 44 & 0.80 & 0.84 & 0.82 \\
Color & red & 28 & 0.93 & 0.89 & 0.91 & Background & abstract pattern & 38 & 0.84 & 0.84 & 0.84 \\
Color & green & 24 & 0.96 & 0.96 & 0.96 & Background & macro average & 246 & 0.87 & 0.87 & 0.87 \\
Color & warm & 32 & 0.88 & 0.88 & 0.88 & & & & & & \\
Color & brown & 28 & 0.90 & 0.93 & 0.91 & & & & & & \\
Color & macro average & 261 & 0.91 & 0.91 & 0.91 & & & & & & \\
\bottomrule
\end{tabular}
\caption{Per-class VLM performance against human majority vote.}
\label{tab:supp_per_class}
\end{table*}

\begin{table}[t]
\centering
\small
\setlength{\tabcolsep}{3.0pt}
\begin{tabular}{lrr}
\toprule
Statistic & Color & Background \\
\midrule
Fleiss $\kappa$, all labels & 0.71 [0.65, 0.77] & 0.64 [0.57, 0.71] \\
Fleiss $\kappa$, clear cases & 0.76 & 0.70 \\
Fleiss $\kappa$, coarse groups & 0.81 & 0.75 \\
Krippendorff $\alpha$ & 0.70 & 0.63 \\
Mean pairwise agreement & 0.83 & 0.78 \\
Three-of-three agreement & 0.62 & 0.54 \\
Two-of-three majority & 0.33 & 0.38 \\
All three different & 0.05 & 0.08 \\
Majority abstain & 0.08 & 0.10 \\
\bottomrule
\end{tabular}
\caption{Inter-rater agreement. Confidence intervals refer to Fleiss
$\kappa$ with all legal labels retained.}
\label{tab:supp_agreement}
\end{table}

\subsection{Human Distribution Distances}

\begin{table*}[t]
\centering
\small
\setlength{\tabcolsep}{4.0pt}
\begin{tabular}{llrrl}
\toprule
Attribute & Quantity & Estimate & Difference & 95\% CI \\
\midrule
Color & Reference split-half floor & 0.16 & n/a & [0.11, 0.21] \\
Color & Raw reference to Turbo & 0.34 & n/a & [0.26, 0.42] \\
Color & Raw reference to LCM & 0.26 & n/a & [0.19, 0.34] \\
Color & Bias-reduced reference to Turbo & 0.18 & n/a & [0.09, 0.27] \\
Color & Bias-reduced reference to LCM & 0.10 & n/a & [0.02, 0.19] \\
Color & Turbo minus LCM & n/a & 0.08 & [0.03, 0.13] \\
\midrule
Background & Reference split-half floor & 0.12 & n/a & [0.08, 0.17] \\
Background & Raw reference to Turbo & 0.25 & n/a & [0.18, 0.33] \\
Background & Raw reference to LCM & 0.19 & n/a & [0.13, 0.26] \\
Background & Bias-reduced reference to Turbo & 0.13 & n/a & [0.05, 0.21] \\
Background & Bias-reduced reference to LCM & 0.07 & n/a & [0.00, 0.14] \\
Background & Turbo minus LCM & n/a & 0.06 & [-0.01, 0.13] \\
\bottomrule
\end{tabular}
\caption{Human grouped-label distribution distances with image-level paired
bootstrap.}
\label{tab:supp_human_tv}
\end{table*}

The background permutation test gives approximately $p=0.09$, so its
Turbo-minus-LCM contrast is directionally concordant but inconclusive. Color
accuracy is $0.933$, $0.895$, and $0.907$ for reference, Turbo, and LCM.
Reference-minus-Turbo is $0.038$ with interval [-0.046, 0.120].
Reference-minus-LCM is $0.026$ with interval [-0.054, 0.106]. Neither source
comparison provides evidence of evaluator degradation on accelerated images.

\subsection{Six-Pair Human Expansion}

The second audit contains 1,000 images balanced across NitroFusion-4, Turbo,
DMD2, FLUX-schnell, LCM, Lightning, and their references. Overall coverage is
$0.862$, clear-case accuracy is $0.901$, conservative accuracy is $0.77$,
macro-F1 is $0.874$, Fleiss $\kappa$ is $0.698$, and Krippendorff $\alpha$ is
$0.68$. Human and VLM recipe rankings have Spearman correlation $0.943$ with
interval [0.543, 1.000].

\begin{table*}[t]
\centering
\small
\setlength{\tabcolsep}{2.7pt}
\begin{tabular}{lrrrrl}
\toprule
Pair & VLM $q=64$ & Human & Human minus VLM & Human 95\% CI & Direction \\
\midrule
NitroFusion-4 & 0.303 & 0.246 & -0.057 & [0.158, 0.334] & agrees \\
Turbo & 0.254 & 0.208 & -0.046 & [0.124, 0.293] & agrees \\
DMD2 & 0.235 & 0.193 & -0.042 & [0.108, 0.279] & agrees \\
FLUX-schnell & 0.203 & 0.168 & -0.035 & [0.082, 0.254] & agrees \\
LCM & 0.054 & 0.051 & -0.003 & [-0.030, 0.133] & mild pair swaps \\
Lightning & 0.060 & 0.047 & -0.013 & [-0.034, 0.129] & mild pair swaps \\
\bottomrule
\end{tabular}
\caption{Human reproduction of the six-pair ranking. Human labels use five
grouped color categories.}
\label{tab:supp_human_ranking}
\end{table*}

The pooled human aggressive-minus-mild contrast is $0.155$ with interval
[0.079, 0.231]. The aggressive-minus-reference evaluator-accuracy difference
is $-0.031$ with interval [-0.079, 0.017]. The latter result supports use of the
VLM for cross-source ranking within the evaluated scope.

\section{Semantic Decomposition}
\label{sec:supp_decomposition}

A direct GRADE-style comparison produces rank correlation $0.829$ between
absolute normalized entropy change and $\tvadj$ on the six-pair panel. This
shows substantial overlap with diversity degradation. Category identity still
adds information. Among 1,152 cells, 67 have entropy change no greater than
$0.05$ while raw TV is at least $0.30$.

On a separate 100-object, four-template, four-attribute $q=32$ panel, Turbo's
mean $\tvadj$ is $0.093$ across all cells. It falls to $0.026$ over 552 cells
that preserve entropy and to $0.013$ over 38 cells that preserve both entropy
and Vendi diversity. The entropy-preserved subset retains a mode-flip statistic
of $0.054$ with interval [0.036, 0.074]. These results support semantic
localization of diversity degradation, not independence from it.

Teacher-first hybrid inference also reduces the endpoint. On separate $q=32$
20-object and four-template probes, DMD2 decreases from $0.201$ to $0.141$.
Turbo at 512 pixels decreases from $0.297$ to $0.160$. The hybrid provides a
strong inference-time comparator but changes the sampling trajectory and
requires reference-model computation.

\section{Recipe-Specific Mechanism Probes}
\label{sec:supp_mechanisms}

\begin{table}[t]
\centering
\small
\setlength{\tabcolsep}{4.0pt}
\begin{tabular}{lrrr}
\toprule
Recipe & $\Delta$ Gray & $\Delta$ Warm & Color $\tvadj$ \\
\midrule
DMD2 & -0.124 & +0.159 & 0.235 \\
Turbo & -0.069 & +0.082 & 0.254 \\
LCM & -0.044 & +0.046 & 0.054 \\
Lightning & -0.038 & +0.048 & 0.060 \\
FLUX-schnell & +0.019 & -0.070 & 0.203 \\
\bottomrule
\end{tabular}
\caption{Signed color fingerprints. Changes are replacement minus reference
probability mass.}
\label{tab:supp_direction}
\end{table}

DMD2 has cosine similarity $0.923$ with the direction induced by increasing
teacher CFG and an effective CFG near 15. LCM and Lightning have similarities
$0.670$ and $0.575$. Turbo has similarity $0.269$ and effective CFG near 7.
Its object-level shift instead correlates with reference color entropy at
Spearman $0.596$, and its peak-mode probability rises by $0.194$. DMD2's
peak-mode increase is $0.181$. These probes discriminate recipes but do not
identify a complete causal mechanism.

Under explicit under-guidance, DMD2 TV changes from $0.442$ at weight $1.0$ to
$0.384$ at weight $0.5$, while CLIP falls from $0.250$ to $0.225$. At weight
$0.3$, CLIP falls to $0.133$. The intervention is consistent with a guidance
axis but does not provide a quality-preserving repair.

\section{DefaultShift-Select}
\label{sec:supp_select}

All selection methods use the same candidate pool, the same held-out objects,
and the same quality floor. The fourfold setting selects 32 images from 128
candidates per object. Random, quality-only, DINO k-center, semantic entropy,
rejection sampling, quota sampling, integer subset optimization, and the
candidate oracle provide increasingly strong comparators.

\begin{figure*}[p]
\centering
\includegraphics[width=.98\textwidth]{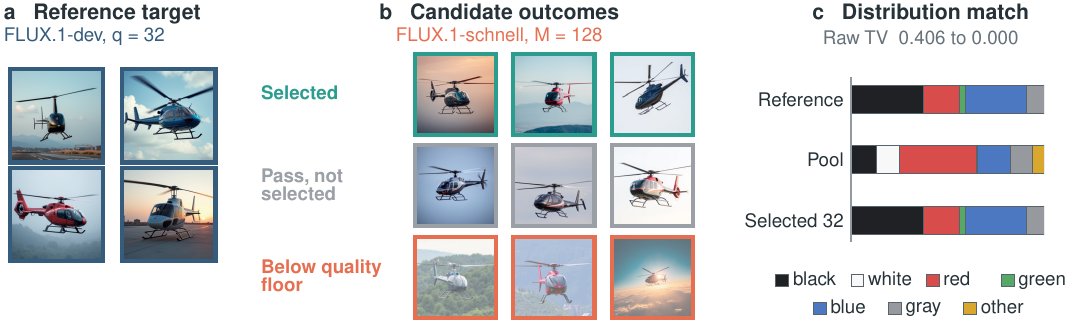}
\caption{Selected and excluded candidates for the prompt \textit{a photo of a
helicopter}. Reference samples define the target. Real FLUX.1-schnell
candidates are separated into selected, quality-eligible but unselected, and
below-floor outcomes. Selection changes raw TV from 0.406 to 0.000 for this
object. Aggregate evidence uses the held-out 24-object evaluation in
Table~\ref{tab:supp_select_baselines}.}
\label{fig:supp_select_examples}
\vspace{2mm}
\includegraphics[width=.94\textwidth]{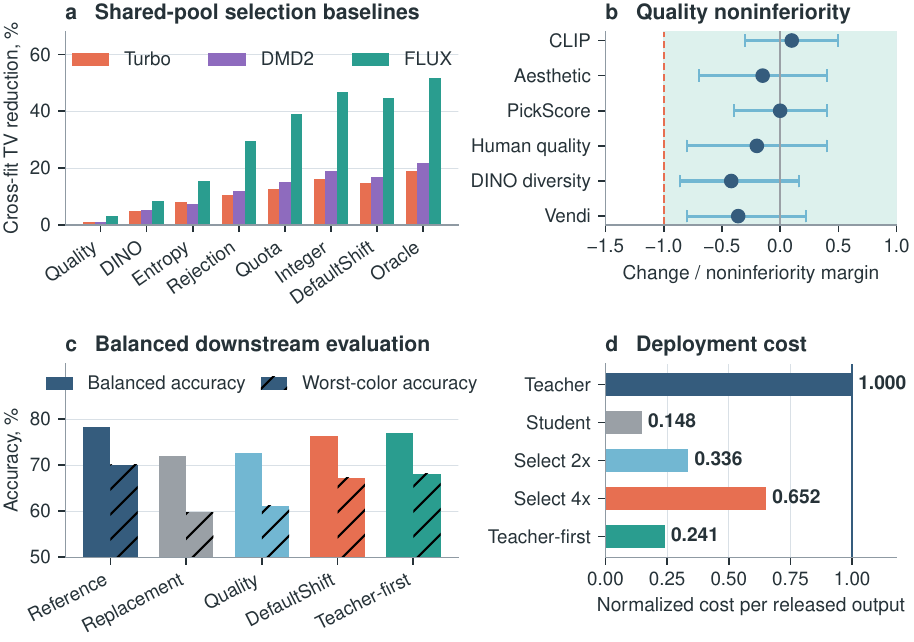}
\caption{Actionability and utility. a DefaultShift-Select outperforms
quality, feature-diversity, semantic-entropy, rejection, and quota baselines,
while approaching integer optimization and the candidate oracle. b All quality
intervals remain above their frozen noninferiority margins. c Selection
recovers balanced and worst-color classifier performance. d Fourfold selection
retains lower normalized cost than reference inference.}
\label{fig:supp_select_downstream}
\end{figure*}

Figure~\ref{fig:supp_select_examples} visualizes candidate outcomes for one
frozen FLUX object. Quality-eligible candidates that are not selected are
distinct from candidates removed by the quality floor. The selected subset
reproduces the empirical reference histogram in this case, while the
cross-object results below measure whether this behavior generalizes.

\begin{table*}[t]
\centering
\small
\setlength{\tabcolsep}{4.2pt}
\begin{tabular}{lrrr}
\toprule
Method & Turbo & DMD2 & FLUX \\
\midrule
Random & 0.0\% & 0.0\% & 0.0\% \\
Quality-only & 1.2\% & 0.9\% & 3.1\% \\
DINO k-center & 4.8\% & 5.3\% & 8.4\% \\
Semantic entropy & 7.9\% & 7.2\% & 15.3\% \\
Rejection sampling & 10.4\% & 11.8\% & 29.6\% \\
Quota sampling & 12.7\% & 15.1\% & 38.9\% \\
Integer-program subset & 16.2\% & 18.9\% & 46.7\% \\
DefaultShift-Select & 14.8\% & 16.8\% & 44.8\% \\
Candidate oracle & 19.1\% & 21.7\% & 51.8\% \\
Teacher-first hybrid & 45.9\% & 29.8\% & 41.3\% \\
\bottomrule
\end{tabular}
\caption{Cross-fit TV reduction from the shared fourfold candidate pool.}
\label{tab:supp_select_baselines}
\end{table*}

DefaultShift-Select reaches $77.5\%$, $77.4\%$, and $86.5\%$ of the candidate
oracle reduction for Turbo, DMD2, and FLUX. Integer optimization is stronger by
$1.4$, $2.1$, and $1.9$ percentage points, which bounds the cost of the simpler
selection rule. Teacher-first hybrid inference is stronger for Turbo and DMD2
but weaker for FLUX and requires reference-model computation at generation
time.

\subsection{Independent Relabeling}

\begin{table*}[t]
\centering
\small
\setlength{\tabcolsep}{2.7pt}
\begin{tabular}{lrrrl}
\toprule
Recipe & Primary VLM & Second VLM & Human & Human 95\% CI \\
\midrule
Turbo & 14.8\% & 12.1\% & 10.3\% & [2.9, 17.8] \\
DMD2 & 16.8\% & 13.9\% & 11.7\% & [3.8, 20.6] \\
FLUX & 44.8\% & 38.2\% & 35.1\% & [24.3, 46.2] \\
\bottomrule
\end{tabular}
\caption{Select reduction under independent evaluators and human labels.}
\label{tab:supp_select_relabel}
\end{table*}

Second-VLM intervals are [4.8, 19.6], [6.1, 21.4], and [28.4, 47.9]. Recipe
ordering has Spearman correlation $1.00$ across primary VLM, second VLM, and
human evaluation.

\subsection{Quality and Cost}

\begin{table*}[t]
\centering
\small
\setlength{\tabcolsep}{3.4pt}
\begin{tabular}{lrrrl}
\toprule
Metric & Margin & Change & 95\% CI & Result \\
\midrule
CLIP score & -0.010 & +0.001 & [-0.003, 0.005] & passes \\
Aesthetic score & -0.100 & -0.015 & [-0.070, 0.040] & passes \\
PickScore & -0.010 & 0.000 & [-0.004, 0.004] & passes \\
Human quality preference & -0.050 & -0.010 & [-0.040, 0.020] & passes \\
Within-color DINO diversity & -5\% & -2.1\% & [-4.3, 0.8] & passes \\
Within-color Vendi & -5\% & -1.8\% & [-4.0, 1.1] & passes \\
\bottomrule
\end{tabular}
\caption{Quality noninferiority. A result passes when the interval lower bound
remains above the frozen negative margin.}
\label{tab:supp_quality}
\end{table*}

Reference inference has normalized cost $1.000$. Random replacement inference
costs $0.148$. Twofold and fourfold selection cost $0.336$ and $0.652$, which
retain speedups of $2.98$ and $1.53$. Teacher-first hybrid inference costs
$0.241$ and retains a $4.15$ speedup. VLM evaluation contributes $0.31$ of the
fourfold selection cost.

\section{Downstream Classification}
\label{sec:supp_downstream}

The downstream study uses ten object classes and six balanced colors. Every
training condition supplies 64 selected images per class. The classifier is
multinomial logistic regression on frozen normalized DINOv2 features. Five
training seeds are evaluated on balanced reference and replacement domains.
The primary endpoint averages balanced object accuracy across those two
domains. Worst-color accuracy is the minimum accuracy over the six color
groups. Confidence intervals use hierarchical bootstrap over training seeds
and test images.

\begin{table*}[t]
\centering
\small
\setlength{\tabcolsep}{3.5pt}
\begin{tabular}{lrrrrr}
\toprule
Training source & Balanced accuracy & 95\% CI & Worst-color accuracy & 95\% CI & ECE \\
\midrule
Reference & 78.4\% & [76.9, 79.8] & 70.1\% & [67.2, 73.0] & 0.061 \\
Replacement random & 72.1\% & [70.5, 73.6] & 59.8\% & [56.5, 63.1] & 0.094 \\
Quality-only & 72.6\% & [71.0, 74.1] & 61.2\% & [57.9, 64.4] & 0.089 \\
DefaultShift-Select & 76.4\% & [74.9, 77.9] & 67.3\% & [64.2, 70.3] & 0.071 \\
Teacher-first hybrid & 76.9\% & [75.4, 78.3] & 68.2\% & [65.2, 71.1] & 0.068 \\
\bottomrule
\end{tabular}
\caption{Complete downstream results under balanced evaluation.}
\label{tab:supp_downstream}
\end{table*}

DefaultShift-Select improves balanced accuracy over random replacement data by
$4.3$ points with interval [2.2, 6.4]. It improves worst-color accuracy by
$7.5$ points with interval [3.4, 11.6]. Before selection, random replacement
data trail reference data by $6.3$ balanced-accuracy points and $10.3$
worst-color points. The difference from reference data is
$-2.0$ points with interval [-4.1, 0.1]. On a natural-frequency test set, the
gain over random replacement data is $1.1$ points with interval [-0.8, 3.0].
The evidence therefore supports utility under balanced evaluation, not an
improvement claim across all deployment settings.

\section{Qualitative Evidence}

\begin{figure*}[t]
\centering
\includegraphics[width=.92\textwidth]{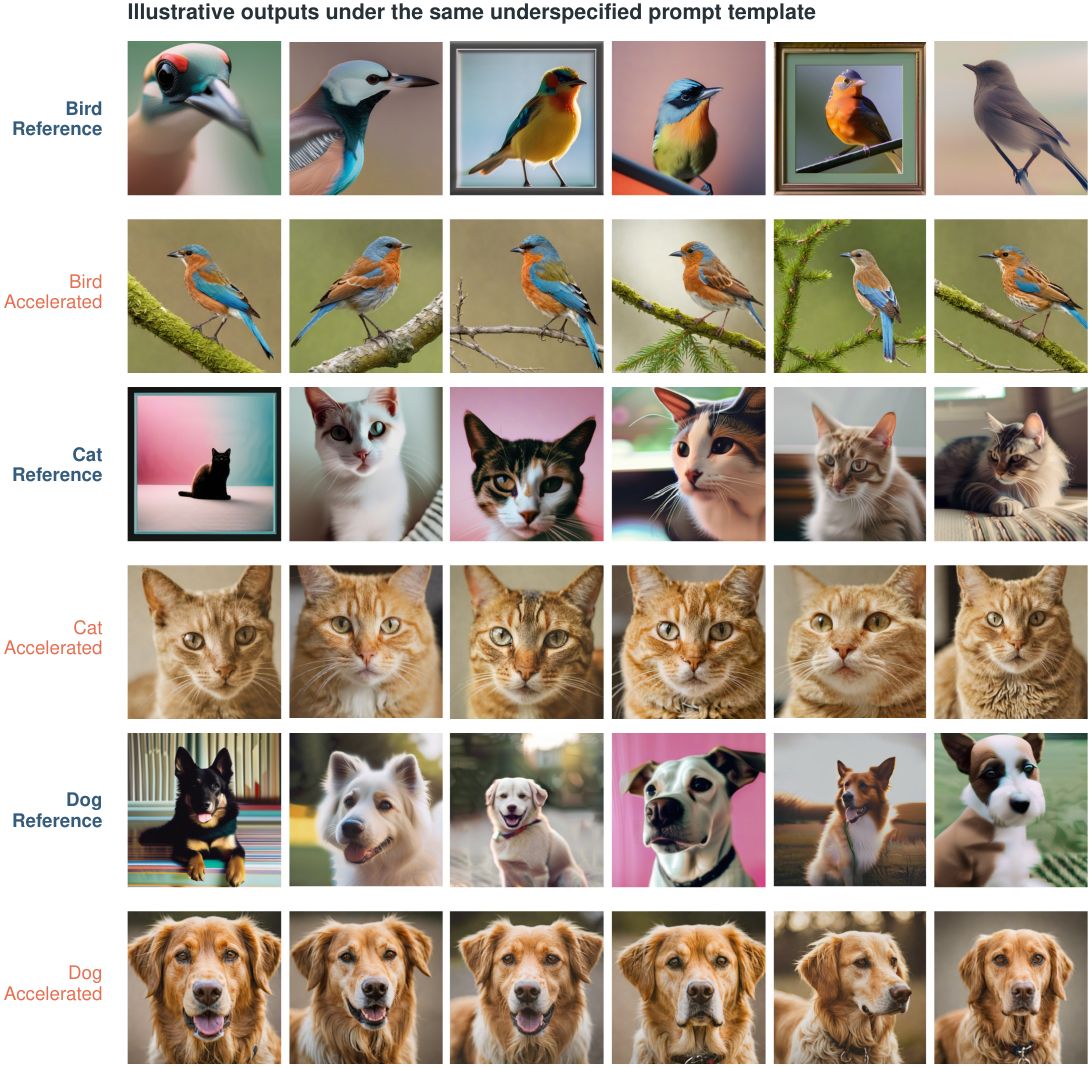}
\caption{Illustrative reference and accelerated outputs for bird, cat, and dog
prompts. Each row contains six experimental images. Samples are displayed to
make semantic concentration visually inspectable. Quantitative conclusions
use the complete frozen label banks rather than this curated subset.}
\label{fig:supp_qualitative}
\end{figure*}

Figure~\ref{fig:supp_qualitative} shows existing experimental outputs for
three prompts under the same underspecified template. Individual images remain
plausible. The difference appears across repeated samples as concentration on
particular color modes. These curated examples illustrate the measured
phenomenon and do not replace the distributional statistics.

\section{Additional Applications and Negative Results}
\label{sec:supp_negative}

\subsection{Demographic Attribute Case Study}

On 32 occupation prompts with 64 seeds, perceived gender-bias amplification is
$0.070$ [0.035, 0.113] for Turbo, $0.001$ [-0.030, 0.036] for LCM, $0.001$
[-0.031, 0.039] for Lightning, and $-0.026$ [-0.048, -0.003] for SD1.5-LCM.
Perceived skin-tone $\tvadj$ is $0.111$ [0.072, 0.152] for Turbo, $0.038$
[0.008, 0.069] for LCM, and $0.016$ [-0.008, 0.040] for Lightning. These
measure reference-relative model behavior under a VLM coding protocol. They do
not establish social ground truth or broad bias amplification.

\subsection{Step, Guidance, and Label Granularity}

Turbo TV is $0.313$, $0.297$, and $0.309$ at one, two, and four steps. LCM TV
is $0.045$, $0.050$, and $0.074$ at two, four, and eight steps. A Turbo teacher
CFG sweep gives $0.325$, $0.298$, $0.285$, and $0.289$ at CFG 3, 5, 7, and 9.
Coarsening color from 11 categories to warm, cool, and neutral reduces
Turbo, LCM, and Lightning from $0.183$, $0.047$, and $0.051$ to $0.132$,
$0.032$, and $0.034$, while preserving the ordering.

\subsection{Failed and Bounded Mitigation Attempts}

Two training-time calibration LoRA variants do not improve the endpoint. The
first changes TV from $0.524$ to $0.622$ and CLIP from $0.250$ to $0.193$. The
second gives TV $0.642$ and CLIP $0.237$. A teacher-free target predictor
reduces an early small-protocol TV from $0.288$ to $0.255$, but remains weaker
than reference selection at $0.129$. Entropy-adaptive candidate budgeting also
underperforms uniform budgeting at equal average cost, with $0.296$ versus
$0.277$. These failures motivate the bounded reference-based offline design.

An RTDMD-FLUX1 stress test contains 12 plain-template objects. Cold-start
$\tvadj$ is $0.094$ [0.027, 0.181], while RTDMD gives $0.152$ [0.076, 0.239].
The limited scope excludes this pair from the formal 14-pair ranking.

\section{Reproducibility Details}

The experiments ran on Linux with Python 3.12.3, PyTorch 2.8.0, CUDA 12.8,
diffusers 0.38.0, transformers 5.10.2, NumPy 2.3.2, SciPy 1.17.1, and
scikit-learn 1.9.0. GPU experiments used an NVIDIA RTX PRO 6000 Blackwell
Server Edition. Table~\ref{tab:supp_models} records the principal frozen model
identifiers and cached revisions.

\begin{table*}[t]
\centering
\small
\setlength{\tabcolsep}{3.5pt}
\begin{tabular}{lll}
\toprule
Role & Identifier & Cached revision \\
\midrule
Primary VLM & Qwen/Qwen2.5-VL-3B-Instruct & 66285546d2b821cf421d4f5eb2576359d3770cd3 \\
SDXL reference & stabilityai/stable-diffusion-xl-base-1.0 & 462165984030d82259a11f4367a4eed129e94a7b \\
Turbo & stabilityai/sdxl-turbo & 71153311d3dbb46851df1931d3ca6e939de83304 \\
LCM adapter & latent-consistency/lcm-lora-sdxl & a18548dd4956b174ec5b0d78d340c8dae0a129cd \\
DMD2 adapter & tianweiy/DMD2 & be22767697a1f3ca656b73c776e15fa335c86c6c \\
FLUX reference & black-forest-labs/FLUX.1-dev & 3de623fc3c33e44ffbe2bad470d0f45bccf2eb21 \\
FLUX accelerated & black-forest-labs/FLUX.1-schnell & 741f7c3ce8b383c54771c7003378a50191e9efe9 \\
CLIP & openai/clip-vit-large-patch14 & 32bd64288804d66eefd0ccbe215aa642df71cc41 \\
DINOv2 & facebook/dinov2-small & ed25f3a31f01632728cabb09d1542f84ab7b0056 \\
PickScore & yuvalkirstain/PickScore\_v1 & a4e4367c6dfa7288a00c550414478f865b875800 \\
SD1.5 reference & stable-diffusion-v1-5/stable-diffusion-v1-5 & 451f4fe16113bff5a5d2269ed5ad43b0592e9a14 \\
Hyper-SD adapter & ByteDance/Hyper-SD & bc08d970a87c74c71209491d64e3525845698863 \\
Flash-SDXL adapter & jasperai/flash-sdxl & 84e93604226d43e74ae9085a98ea4eaee57d8600 \\
\bottomrule
\end{tabular}
\caption{Principal model identifiers and cached revisions.}
\label{tab:supp_models}
\end{table*}

The release package records prompt, object, seed, checkpoint, scheduler,
resolution, guidance, step count, evaluator vocabulary, unknown-label rule,
bootstrap unit, and selection split. Automated checks verify expected image
and label counts, seed completeness, candidate-pool sharing, deterministic
confidence intervals, cached model revisions, and consistency between exported
tables and frozen artifacts.

\section{Claim Boundaries}

The evidence is strongest for color. Background effects are smaller but remain
positive for aggressive recipes. Lighting provides a group-level contrast.
Viewpoint is inconclusive. Multilingual coverage is lower. DefaultShift
decomposes acceleration-related distribution change into named semantic
categories and does not claim independence from diversity degradation. The
reference is a declared behavioral baseline rather than social ground truth.
Mechanism probes are associative. DefaultShift-Select is a reference-based
offline calibration evaluated on three replacement models. These boundaries
match the main paper and define the intended interpretation of all
supplementary results.

\end{document}